\pdfoutput=1
\RequirePackage{fix-cm}

\documentclass[twocolumn]{svjour3}          
\smartqed  
\usepackage{graphicx}
\usepackage{verbatim} 
\usepackage{amssymb}
\usepackage{amsmath}
\usepackage{hyphenat}
\usepackage{algorithmic}
\usepackage{subfigure}
\usepackage{wasysym}
\usepackage{universal}
\usepackage{color}

\journalname{IJCV}
\begin{document}

\title{Hypothesize and Bound: A Computational Focus of Attention Mechanism for Simultaneous $N$-D Segmentation, Pose Estimation and Classification Using Shape Priors}



\author{Diego Rother \and Simon Sch\"utz \and Ren\'e Vidal}   


\institute{D. Rother \and R. Vidal \at
              Johns Hopkins University \\
              Tel.: +1-410-516-6736\\
              \email{diroth@gmail.com}           
           \and
           S. Sch\"utz \at
           University of G\"ottingen
}

\date{Received: date / Accepted: date}

\maketitle

\begin{abstract}
Given the ever increasing bandwidth of the visual sensory information available to autonomous a\-gents and other automatic systems, it is becoming essential to endow them with a sense of what is worthwhile their attention and what can be safely disregarded. This article presents a general mathematical framework to efficiently allocate the available computational resources to process the parts of the input that are relevant to solve a perceptual problem of interest. By solving a perceptual problem we mean to find the hypothesis $H$ (i.e., the state of the world) that maximizes a function $L(H)$, referred to as the evidence, representing how well each hypothesis ``explains'' the input. However, given the large bandwidth of the sensory input, fully evaluating the evidence for each hypothesis is computationally infeasible (e.g., because it would imply checking a large number of pixels). To address this problem we propose a mathematical framework with two key ingredients. The first one is a Bounding Mechanism (BM) to compute lower and upper bounds of the evidence of a hypothesis, for a given computational budget. These bounds are much cheaper to compute than the evidence itself, can be refined at any time by increasing the budget allocated to a hypothesis, and are frequently sufficient to discard a hypothesis. The second ingredient is a Focus of Attention Mechanism (FoAM) to select which hypothesis' bounds should be refined next, with the goal of discarding non-optimal hypotheses with the least amount of computation.

The proposed framework has the following desirable characteristics: 1) it is very efficient since most hypotheses are discarded with minimal computation; 2) it is parallelizable; 3) it is guaranteed to find the globally optimal hypothesis or hypotheses; and 4) its running time depends on the problem at hand, not on the bandwidth of the input. In order to illustrate the general framework, in this article we instantiate it for the problem of simultaneously estimating the class, pose and a noiseless version of a 2D shape in a 2D image. To do this, we develop a novel theory of semidiscrete shapes that allows us to compute the bounds required by the BM. We believe that the theory presented in this article (i.e., the algorithmic paradigm and the theory of shapes) has multiple potential applications well beyond the application demonstrated in this article.

\keywords{Focus of Attention  \and shapes \and shape priors \and hypothesize-and-verify \and coarse-to-fine \and probabilistic inference \and graphical models \and image understanding}

\end{abstract}

\section{Introduction}
\label{sec:Introduction}
\begin{figure}
\includegraphics[width=\columnwidth, bb=0pt 0pt 650pt 184pt]{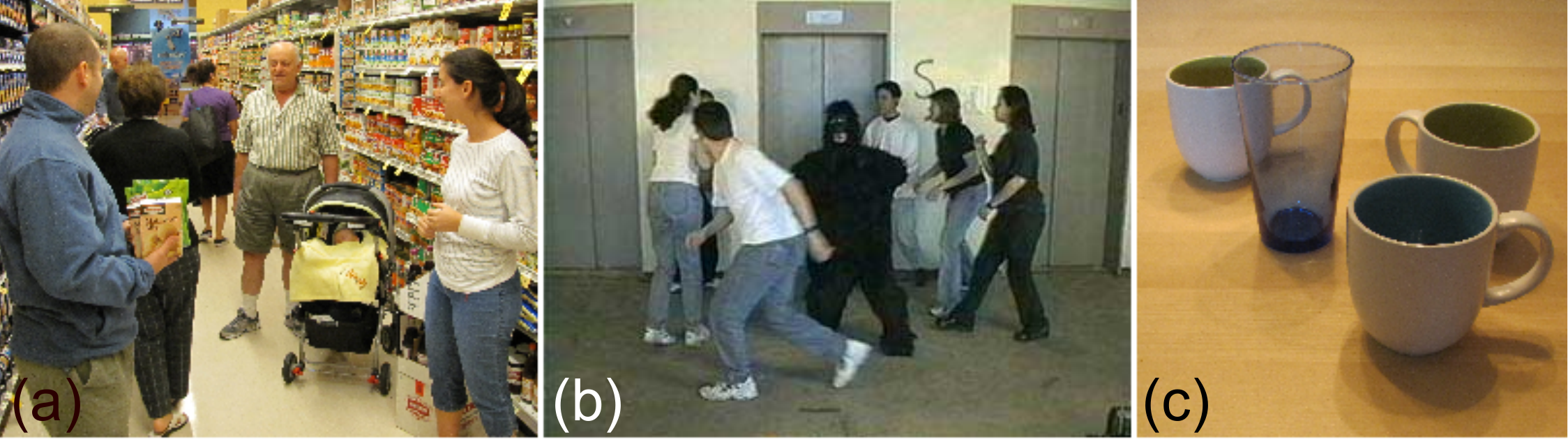}
\caption{(a) Natural scenes are so complex that it would be computationally wasteful for humans to extract \emph{all} the information that they can extract from them. (b) Attention is task oriented: when focused on the task of counting the passes of a ball between the people in this video, many humans fail to see the gorilla in the midle of the frame (image from \cite{Sim99}, see details in that article). (c) Precision is task dependent: in order to pick up the glass, one does not need to estimate the locations of the mugs and the glass with the same precision. A rough estimate of the locations of the mugs is enough to avoid them; a better estimate is needed to pick up the glass.}
\vspace{-10pt}
\label{fig:FoAMMotivation}
\end{figure}

Humans are extremely good at extracting information from images. They can recognize objects from many different classes, even objects they have never seen before; they can estimate the relative size and position of objects in 3D, even from 2D images; and they can (in general) do this in poor lighting conditions, in cluttered scenes, or when the objects are partially occluded. However, natural scenes are in general very complex (Fig. \ref{fig:FoAMMotivation}a),  full of objects with intricate shapes, and so rich in textures, shades and details, that extracting \emph{all} this information would be computationally very expensive, even for humans. Experimental psychology studies, on the other hand, suggest that (mental) computation is a limited resource that, when demanded by one task, is unavailable for another \cite{Kah73}. This is arguably why humans have evolved a \emph{focus of attention mechanism} (FoAM) to discriminate between the information that is needed to achieve a specific goal, and the information that can be safely disregarded.

Humans, for example, do not perceive \emph{every} object that enters their field of view; rather they perceive \emph{only} the objects that receive their focused attention (Fig. \ref{fig:FoAMMotivation}b) \cite{Sim99}. Moreover, in order to save computation, it is reasonable that even for objects that are indeed perceived, only the details that are relevant towards a specific goal are extracted (as beautifully illustrated in \cite{dot08}). In particular, it is reasonable to think that objects are not \emph{classified} at a more concrete level than necessary (e.g., `terrier' vs. `dog'), if this were more expensive than classifying the object at a more abstract level, and if this provided the same amount of relevant information towards the goal \cite{Ros78}. Also we do not expect computation to be spent estimating other properties of an object (such as \emph{size} or \emph{position}) with higher precision than necessary if this were more expensive and provided the same amount of relevant information towards the goal (Fig. \ref{fig:FoAMMotivation}c).

\subsection{Hypothesize-and-bound algorithms}
\label{sec:SCAlgorithms}
In this article we propose a mathematical framework that uses a FoAM to allocate the available computational resources where they contribute the most to solve a given task. This mechanism is one of the parts of a novel family of inference algorithms that we refer to as \emph{hypothesize-and-bound} (H\&B) algorithms. These algorithms are based on the \emph{hypothesize-and-verify} paradigm. In this paradigm a set of \emph{hypotheses} and a function referred to as the \emph{evidence} are defined. Each hypothesis $H$ represents a different state of the world (e.g., which objects are located where) and its evidence $L(H)$ quantifies how well this hypothesis ``explains'' the input image. In a typical hypothesize-and-verify algorithm the evidence of each hypothesis is evaluated and the hypothesis (or group of hypotheses) with the highest evidence is selected as the optimal. 

However, since the number of hypotheses could be very large, it is essential to be able to evaluate the evidence of each hypothesis with the least amount of computation. For this purpose the second part of a H\&B algorithm is a \emph{bounding mechanism} (BM), which computes lower and upper bounds for the evidence of a hypothesis, instead of evaluating it exactly. These bounds are in general significantly less expensive to compute than the evidence itself, and they are obtained for a given computational budget (allocated by the FoAM), which in turn defines the tightness of the bounds (i.e., higher budgets result in tighter bounds). In some cases, these inexpensive bounds are already sufficient to discard a hypothesis (e.g., if the upper bound of $L(H_1)$ is lower than the lower bound of $L(H_2)$, $H_1$ can be safely discarded). Otherwise, these bounds can be efficiently and progressively refined by spending extra computational cycles on them (see Fig. \ref{fig:EvolutionOfBounds}). As mentioned above, the Fo\-AM allocates the computational budget among the different hypotheses. This mechanism keeps track of the progress made (i.e., how much the bounds got closer to each other) for each computation cycle spent on each hypothesis, and decides on-the-fly where to spend new computation cycles in order to economically discard as many hypotheses as possible. Because computation is allocated where it is most needed, H\&B algorithms are in general \emph{very efficient}; and because hypotheses are discarded only when they are \emph{proved} suboptimal, H\&B algorithms are \emph{guaranteed to find the optimal solution}. 

\subsection{Application of H\&B algorithms to vision}
\label{sec:ApplicationSC}
The general inference framework mentioned in the previous paragraphs is applicable to any problem in whi\-ch bounds for the evidence can be inexpensively obtained for each hypothesis. Thus, to instantiate the framework to solve a particular problem, a specific BM has to be developed for that problem. (The FoAM, on the other hand, is common to many problems since it only communicates with the BM by allocating the computational budget to the different hypotheses and by ``reading" the resulting bounds.) 

In this article we illustrate the framework by instantiating it for the specific problem of jointly estimating the \emph{class} and the \emph{pose} of a 2D shape in a noisy 2D image, as well as recovering a noiseless version of this shape. This problem is solved by ``merging" information from the input image and from probabilistic models known \emph{a priori} for the shapes of different classes.

As mentioned above, to instantiate the framework for this problem, we must define a mechanism to compute and refine bounds for the evidence of the hypotheses. To do this, we will introduce a novel theory of shapes and shape priors that will allow us to efficiently compute and refine these bounds. While still of practical interest, we believe that the problem chosen is simple enough to best illustrate the main characteristics of H\&B algorithms and the proposed theory of shapes, without occluding its main ideas. Another instantiation of H\&B algorithms, which we describe in \cite{Rot102}, tackles the more complex problem of \emph{simultaneous object classification, pose estimation, and 3D reconstruction, from a single 2D image}. In that work we also use the theory of shapes and shape priors to be described in this article to construct the BM for that problem. 

\subsection{Paper contributions}
\label{sec:Contributions}
The framework we propose has several novel contributions that we group in two main areas, namely: 1) the inference framework using H\&B algorithms; and 2) the shape representations, priors, and the theory developed around them. The following paragraphs summarize these contributions, while in the next section we put them in the context of prior work.

The first contribution of this paper is the use of H\&B algorithms for inference in probabilistic graphical models. In particular, inference in graphical models containing loops. This inference method is not general, i.e., it is not applicable to \emph{any} directed graph with loops. Rather it is specifically designed for the kinds of graphs containing pixels and voxels that are often used for vision tasks (which typically have a large number of variables and a huge number of loops among these variables). The proposed inference framework has several desirable characteristics. First, it is general, in the sense that the only requirement for its application is a BM to compute and refine bounds for the evidence of a hypothesis. Second, the framework is \emph{computationally very efficient}, because it allocates computation dynamically where it is needed (i.e., refining the most promising hypotheses and examining the most informative image regions). Third, the total computation does not depend on the arbitrary resolution of the input image, but rather on the task at hand, or more precisely, on the similarity between the hypotheses that are to be distinguished. (In other words, ``easy'' tasks are solved very fast, while only ``difficult'' tasks require processing the image completely.) This allows us to avoid the common preprocessing step of downsampling the input image to the maximal resolution that the algorithm can handle, and permits us to use the original (possibly very high) resolution only in the parts of the image where it is needed. Fourth, the framework is \emph{fully parallelizable}, which allows it to take advantage of GPUs or other parallel architectures. And fifth, it is \emph{guaranteed to find the globally optimal solution} (i.e., the hypothesis with the maximum evidence), if it exists, or a set of hypotheses that can be formally proved to be undistinguishable at the maximum available image resolution. This guarantee is particularly attractive when the subjacent graphical model contains many loops, since existing probabilistic inference methods are either very inefficient or not guaranteed to find the optimal solution. 

The second contribution relates to the novel shape representations proposed, and the priors presented to encode the shape knowledge of the different object cla\-sses. These shape representations and priors have three distinctive characteristics. First, they are able to represent a shape with \emph{multiple levels of detail}. The level of detail, in turn, defines the amount of computation required to process a shape, which is critical in our framework. Second, it is \emph{straightforward and efficient to project a 3D shape} expressed in these representations to the 2D image plane. This will be essential in the second part of this work \cite{Rot102} to efficiently compute how well a given 3D reconstruction ``explains'' the input image. And third, based on the theory developed for these shape representations and priors, it is possible to \emph{efficiently compute tight log-probability bounds}. Moreover, the tightness of these bounds also depends on the level of detail selected, allowing us to dynamically trade computation for bound accuracy. In addition, the theory introduced is general and could be applied to solve many other problems as well.

\subsection{Paper organization}
\label{sec:Organization}
The remainder of this paper is organized as follows. Section \ref{sec:PriorWork} places the current work in the context of prior relevant work, discussing important connections. Section \ref{sec:FoAM} describes the proposed FoAM. To illustrate the application of the framework to the concrete problem of 2D shape classification, denoising, and pose estimation, in Section \ref{sec:ProblemDefinition} we formally define this problem, and in Section \ref{sec:BMDirectProblem} we develop the BM for it. In order to develop the BM, we first introduce in Section \ref{sec:TheoryOfShapes} a theory of shapes and shape priors necessary to compute the desired bounds for the evidence. Because the FoAM and the theory of shapes described in section \ref{sec:FoAM} and \ref{sec:TheoryOfShapes}, respectively, are general (i.e., not only limited to solve the problem described in Section \ref{sec:ProblemDefinition}), these sections were written to be self contained. Section \ref{sec:ExperimentalResults} presents experimental results obtained with the proposed framework, and Section \ref{sec:Conclusions} concludes with a discussion of the key contributions and directions for future research. In the continuation of this work \cite{Rot102}, we extend the theory presented in this article to deal with a more complex problem involving not just 2D shapes, but also 3D shapes and their 2D projections.

\section{Prior work}
\label{sec:PriorWork}

As mentioned in the previous section, this article pre\-sents contributions in two main areas: 1) the inference framework based on the FoAM; and 2) the shape representations proposed and the theory developed around them. For this reason, in this section we briefly review prior related work in these areas.

\subsection{FoAM}
\label{sec:PWFoAM}
Many computational approaches that rely on a focus of attention mechanism have been proposed over the years, in particular to interpret \emph{visual} stimuli. These computational approaches can be roughly classified into two groups, depending on whether they are biologically inspired or not. 

Biologically inspired approaches \cite{Fri10,Tso05,Shi07}, by definition, exploit characteristics of a model proposed to describe a biological system. The goals of these approaches often include: 1) to validate a model proposed for a biological system; 2) to attain the outstanding performance of biological systems by exploiting some characteristics of these systems; and 3) to facilitate the interaction between humans and a robot by emulating mechanisms that humans use (e.g., joint attention \cite{Kap06}). 
Biological strategies, though optimized for eyes and brains during millions of years of evolution, \emph{are not necessarily optimal for current cameras and computer architectures}. Moreover, since the biological attentional strategies are adopted at the foundations of these approaches by fiat (instead of emerging as the solution to a formally defined problem), it is often \emph{difficult to rigorously analyze the optimality of these strategies}. In addition, these attentional strategies were in general empirically discovered for a particular sensory modality (predominantly vision) and are \emph{not directly applicable to other sensory modalities}. Moreover, these strategies are not general enough to handle simultaneous stimuli coming from several sensory modalities (with some exceptions, e.g., \cite{Arr09}).

Since in this article we are mainly interested in improving the performance of a perceptual system (possibly spanning several sensory modalities), and since we want to be able to obtain optimality guarantees, we do not focus further on biologically inspired approaches.

The second class of focus of attention mechanisms contains those approaches that are not biologically inspired. Within this class we focus on those approaches that are not \emph{ad hoc} (i.e., they are rigorously derived from first principles) and are general (i.e., they are able to handle different sensory modalities and tasks). This subclass contains at least two other approaches (apart from ours): \emph{Branch and Bound} (B\&B) and \emph{Entropy Pursuit} (EP).

In a B\&B algorithm \cite{Cla97}, as in a H\&B algorithm, an objective function is defined over the hypothesis space and the goal of the algorithm is to select the hypothesis that maximizes this function. A B\&B algorithm proceeds by dividing the hypothesis space into subspaces, computing bounds of the objective function for each subspace (rather than for each hypothesis in the subspace), and discarding subspaces that can be proved to be non-optimal (because their upper bound is lower than that of some other subspace). In these algorithms computation is saved by evaluating whole groups of hypotheses, instead of evaluating each individual hypothesis. In contrast, in our approach the hypothesis space is discrete and bounds are computed for \emph{every hypothesis} in the space. In this case computation is saved by discarding most of these hypotheses with very little computation. In other words, for most hypotheses the inexpensive bounds computed for the evidence of these hypotheses are enough to discard these hypotheses. As will be discussed in Section \ref{sec:Conclusions}, this approach is complementary to B\&B, and it would be beneficial to integrate both approaches into a single framework. Due to space limitations, however, this is not addressed in this article.

In an EP algorithm \cite{Gem96,Szn10} a probability distribution is defined over the hypothesis space. Then, during each iteration of the algorithm, a test is performed on the input, and the probability distribution is updated by taking into account the result of the test. This test is selected as the one that is expected to reduce the entropy of the distribution the most. The algorithm terminates when the entropy of the distribution falls below a certain threshold.
A major difference between EP and B\&B/H\&B algorithms is that in each iteration of EP a \emph{test} is selected and the probability of each (of potentially too many) \emph{hypothesis} is updated. In contrast, in each iteration of B\&B/H\&B, one \emph{hypothesis (or one group of hypotheses)} is selected and only the bounds corresponding to it are updated. Unlike B\&B and H\&B algorithms, EP algorithms are not guaranteed to find the optimal solution.

A second useful criterion to classify computational approaches that rely on a FoAM considers whether attention is controlled only by ``bottom-up'' signals derived from salient stimuli, or whether it is also controlled by ``top-down'' signals derived from task demands, or from what a model predicts to be most relevant. \emph{Bottom-up} approaches (e.g., \cite{Koc85}) are also known as \emph{data-driven} approaches, while \emph{top-down} approaches (e.g., \cite{Oli03}) are also known as \emph{task-driven} approaches. Even though the significance of top-down signals in biological systems is well known, most current computer systems only consider bottom-up signals \cite{Fri10}. In contrast, all the three algorithmic paradigms described (H\&B, B\&B and EP), depending on the specific instantiation of these paradigms, are able to handle bottom-up as well as top-down signals. In particular, in the instantiation of H\&B algorithms presented in Section \ref{sec:ProblemDefinition}, both kinds of signals are considered (in fact, it will be seen in Equation \eqref{eq:L_H} that they play a symmetric role). In addition, in all the three algorithmic paradigms described above there is an explicit FoAM to control where the computation is allocated.

\subsection{Inference framework}
\label{sec:InferenceFramework}
Many methods have been proposed to perform inference in graphical models \cite{Kol09}. \emph{Message passing algorithms} are one class of these methods. \emph{Belief propagation} (BP) is an algorithm in this class that is guaranteed to find the optimal solution in a loopless graph (i.e., a polytree) \cite{Bis06}. The \emph{loopy belief propagation} (LBP) and the \emph{junction tree} (JT) algorithms are two algorithms that extend the capabilities of the basic BP algorithm to handle graphs with loops. In LBP messages are exchanged exactly as in BP, but multiple iterations of the basic BP algorithm are required to converge to a solution. Moreover, the method is not guaranteed to converge to the optimal solution in every graph but only in some types of graphs \cite{Wei01}. In the JT algorithm \cite{Lau88}, BP is run on a modified graph whose cycles have been eliminated. To construct this modified graph, the first step is ``moralization,'' which consists of marrying the parents of all the nodes. For the kinds of graphs we are interested in, however, this dramatically increases the clique size. While the JT algorithm is guaranteed to find the optimal solution, in our case this algorithm is not efficient because its complexity grows exponentially with the size of the largest clique in the modified graph.

Two standard ``tricks'' to perform exact inference (using BP) by eliminating the loops of a general graph are: 1) to merge nodes in the graph into a ``supernode,'' and 2) to make assumptions about the values of (i.e., to instantiate) certain variables, creating a different graph for each possible value of the instantiated variables (i.e., for each hypothesis) \cite{Pea88}. These approaches, however, bring their own difficulties. On the one hand, merging nodes results in a supernode whose number of states is the product of the number of states of the merged nodes. On the other hand, instantiating variables forces us to solve an inference problem for a potentially very large number of hypotheses.

In this work we propose a different approach to merge nodes that does not run into the problems mentioned above. Specifically, instead of assuming that the image domain is composed of a finite number of discrete pixels and merging them into supernodes, we assume that the image domain is continuous and consists of an infinite number of ``pixels.'' We then compute ``summaries'' of the values of the pixels in each region of the domain (this is formally described in Section \ref{sec:TheoryOfShapes}). In order to solve the inference efficiently for each hypothesis, the total computation per hypothesis is trimmed down by using lower and upper bounds and a FoAM, as mentioned in Section \ref{sec:Introduction}.

\subsection{Shape representations and priors}
\label{sec:ShapeRepresentations}
Since shape representations and priors are such essential parts of many vision systems, over the years many shape representations and priors have been proposed (see reviews in \cite{Cos09,Dry98}). Among these, only a small fraction have the three properties required by our system and mentioned in Section \ref{sec:Contributions}, i.e., support multiple levels of detail, efficient projection, and efficient computation of bounds.

Scale space and orthonormal basis representations have the property that they can encode multiple levels of detail. In the \emph{scale-space representation} \cite{Lin94}, a shape (or image in general) is represented as a one-parameter family of smoothed shapes, parameterized by the size of the smoothing~kernel~used for suppressing fine-scale structures. Therefore, the representation contains a smoothed copy of the original shape at each level of detail. In the \emph{orthonormal basis representation}, on the other hand, a shape is represented by its coefficients in an orthonormal basis. To compute these coefficients the shape is first expressed in the same representation as the orthonormal basis. For example, in \cite{Nai05} and \cite{Per86} the contour of a shape is expressed in spherical wavelets and Fourier bases, respectively, and in \cite{San09} the signed distance function of a shape is written in terms of the principal components of the signed distance functions of shapes in the training database. The level of detail in this case is defined by the number of coefficients used to represent the shape in the basis. While these shape representations have the first property mentioned above (i.e., multiple levels of detail), they do not have the other two, that is that it is not trivial to efficiently project 3D shapes expressed in these representations to the 2D image plane, or to compute the bounds that we want to compute.

The shape representations we propose, referred to as discrete and semidiscrete shape representations (defined in Section \ref{sec:TheoryOfShapes} and shown in Fig. \ref{fig:ShapeRepresentations}) are respectively closer to region quadtrees/octrees \cite{Sam88} and to occupancy grids \cite{Elf89}. In fact, the discrete shape representation we propose is a special case of a \emph{region quadtree/octree} in which the rule to split an element is a complex function of the input data, the prior knowledge, and the interaction with other hypotheses. Quad\-trees and octrees have been previously used for 3D recognition \cite{Chi86} and 3D reconstruction \cite{Pot87} from multiple silhouettes (not from a single one,  to the best of our knowledge, as we do in \cite{Rot102}). \emph{Occupancy grids}, on the other hand, are significantly different from semidiscrete shapes since they store at each cell a qualitatively different quantity: occupancy grids store the \emph{posterior probability} that an object is in the cell, while semidiscrete shapes store the \emph{measure} of the object in the cell.

\section{Focus of attention mechanism}
\label{sec:FoAM}

In Section \ref{sec:Introduction} we mentioned that a \emph{hypothesize-and-bound} (H\&B) algorithm has two parts: 1) a \emph{focus of attention mechanism} (FoAM) to allocate the available computation cycles among the different hypotheses; and 2) a \emph{bounding mechanism} (BM) to compute and refine the bounds of each hypothesis. In this section we describe in detail the first of these two parts, the FoAM.  

Let $I$ be some input and let ${\mathbb H}=\left\{H_1,\dots ,H_{N_H}\right\}$ be a set of $N_H$ hypotheses proposed to ``explain'' this input. In our problem of interest (formally described in Section \ref{sec:ProblemDefinition}) the input $I$ is an image, and each of the hypotheses corresponds to the \emph{2D pose} and \emph{class} of a 2D shape in this input image. However, from the point of view of the FoAM, it is not important what the input actually is, or what the hypotheses actually represent. The input can be simply thought of as ``some information about the world acquired through some sensors,'' and the hypotheses can be simply thought of as representing a ``possible state of the world.''

Suppose that there exist a function $L(H)$ that quantifies the \emph{evidence} in the input $I$ supporting the hypothesis $H$. In Section \ref{sec:ProblemDefinition} the evidence for our problem is shown to be related to the log-joint probability of the image $I$ and the hypothesis $H$. But again, from the point of view of the FoAM, it is not important how this function is defined; it only matters that hypotheses that ``explain" the input better produce higher values. Thus, part of the goal of the FoAM is to select the hypothesis (or group of hypotheses) $H_{i^*}$ that best explain the input image, i.e.,
\begin{equation}
H_{i^*} = \operatorname*{arg\,max}_{H \in \mathbb H} L(H).
\label{eq:FoAMGoal}
\end{equation}

Now, suppose that the evidence $L(H_i)$ of a hypothesis $H_i$ is very costly to evaluate (e.g., because a large number of pixels must be processed to compute it), but lower and upper bounds for it, $\underline{L}(H_i)$ and $\overline{L}(H_i)$, respectively, can be cheaply computed by the BM. Moreover, suppose that the BM can efficiently refine the bounds of a hypothesis $H_i$ if additional \emph{computational cycles} (defined below) are allocated to the hypothesis. Let us denote by $\underline{L}_{n_i}(H_i)$ and $\overline{L}_{n_i}(H_i)$, respectively,  the lower and upper bounds obtained for $L(H_i)$ after $n_i$ computational cycles have been spent on $H_i$. If the BM is well defined, the bounds it produces must satisfy
\begin{equation}
\underline{L}_{n_i+1}(H_i) \ge \underline{L}_{n_i}(H_i), \ \ and \  \  \ \overline{L}_{n_i+1}(H_i) \le \overline{L}_{n_i}(H_i),
\label{eq:PropertiesBounds}
\end{equation}
for every hypothesis $H_i$, and every $n_i \ge 0$ (assume that $n_i=0$ is the initialization cycle in which the bounds are first computed). In other words, the bounds must not become looser as more computational cycles are invested in their computation. Note that we expect different numbers of cycles to be spent on different hypotheses, ideally with ``bad" hypotheses being discarded earlier than ``better" ones (i.e., $n_1 < n_2$ if $L(H_1) \ll L(H_2)$).

The ``computational cycles'' mentioned above are our unit to measure the computational resources spent. Each \emph{computational cycle}, or just \emph{cycle}, is the computation that the BM \emph{spends} to refine the bounds. While the exact conversion rate between cycles and operations depends on the particular BM used, what is important from the point of view of the FoAM is that all refinement cycles take approximately the same number of operations (defined to be equal to one computational cycle).

The full goal of the FoAM can now be stated as to select the hypothesis $H_i$ that satisfies
\begin{equation}
\underline{L}_{n_i}(H_i) > \overline{L}_{n_j}(H_j)\ \ \forall j \ne i,
\label{eq:TerminationCondition1}
\end{equation}
while minimizing the total number of cycles spent, $\sum^{N_H}_{j=1}$ $n_j$. If these inequalities are satisfied, it can be proved that $H_i$ is the optimal hypothesis, without having to compute \emph{exactly} the evidence for every hypothesis (which is assumed to be much more expensive than just computing the bounds). However, it is possible that after all the hypotheses in a set ${\mathbb H}_i \subset {\mathbb H}$ have been refined to the fullest extent possible, their upper bounds are still bigger than or equal to the maximum lower bound $\gamma \triangleq \mathop{\max }_{H_i \in {\mathbb H}_i} \underline{L}_{n_i}(H_i)$, i.e.,
\begin{equation}
\overline{L}_{n_i}(H_i) \ge \gamma \ \ \ \forall H_i \in \mathbb{H}_i.
\label{eq:TerminationCondition2}
\end{equation}
In this situation all the hypotheses in ${\mathbb H}_i$ \emph{could} possibly be optimal, but we cannot say which one actually \emph{is}. We just do not have the right input to distinguish between them (e.g., because the resolution of the input image is insufficient). We say that these hypotheses are \emph{indistinguishable} given the current input. In short, the FoAM will terminate either because it has \emph{found the optimal hypothesis} (satisfying \eqref{eq:TerminationCondition1}), or because it has \emph{found a set of hypotheses that are indistinguishable from the optimal hypothesis given the current input} (and satisfies \eqref{eq:TerminationCondition2}).

These \emph{termination conditions} can be achieved by very different \emph{budgets} that allocate different number of cycles to each hypothesis. We are interested in finding the budget that achieves them in the minimum number of cycles. Finding this minimum is in general not possible since the FoAM does not ``know,'' \emph{a priori}, how the bounds will change for each cycle it allocates to a hypothesis. For this reason, we propose a heuristic to select the next hypothesis to refine at each point in time. Once a hypothesis is selected, one cycle is allocated to this hypothesis, which is thus refined once by the BM. This selection-refinement cycle is continued until termination conditions are reached.

According to the heuristic proposed, the next hypothesis to refine, $H_{i*}$, is chosen as the one that is expected to produce the greatest reduction $\Delta P(H_{i*})$ in the following potential $P$,
\begin{equation}
P\triangleq \sum_{H_i \in \mathbb{A}}{\left(\overline{L}_{n_i}(H_i) - \gamma \right)},
\label{eq:Potential}
\end{equation}
where $\mathbb{A}$ is the set of all the hypotheses not yet discarded (i.e., those that are \emph{active}), $\gamma$ is the maximum lower bound defined before, and $n_i$ is the number of refinement cycles spent on hypothesis $H_i$. This particular expression for the potential was chosen for two reasons: 1) because it reflects the workload left to be done by the FoAM; and 2) because it is minimal when termination conditions have been achieved. 

\begin{figure}
\vspace{-5pt}
\includegraphics[width=\columnwidth, bb=0pt 0pt 354pt 136pt]{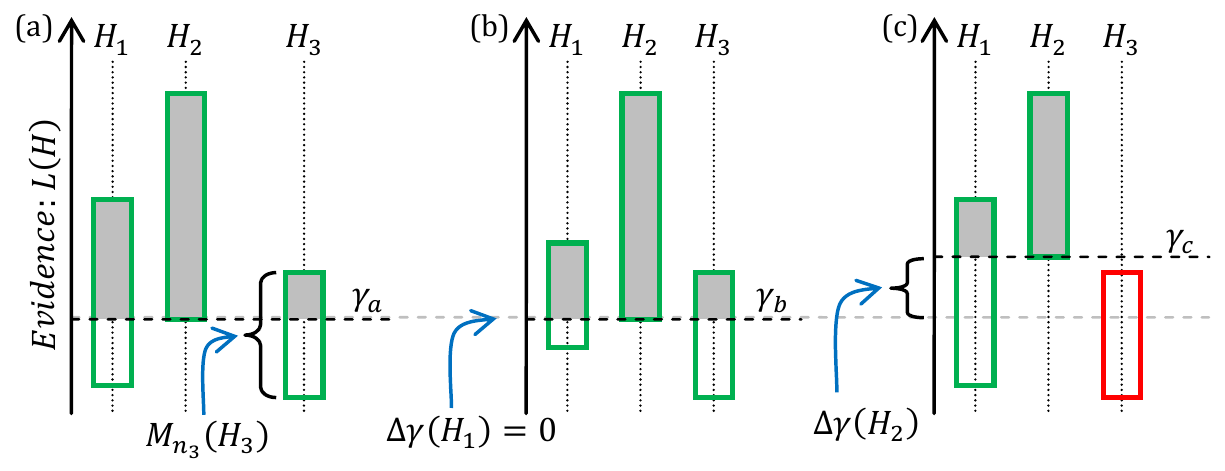}
\vspace{-15pt}
\caption{(a) Bounds for the evidence $L(H_i)$ of three active hypotheses ($H_1$, $H_2$, and $H_3$) after refinement cycle $t$. (b-c) Bounds for the same three hypotheses after refinement cycle $t+1$, assumming that either $H_1$ (b) or $H_2$ (c) was refined during cycle $t+1$. Each rectangle represents the interval where the evidence of a hypothesis is known to be. \emph{Green} and \emph{red} rectangles denote \emph{active} and \emph{discarded} hypotheses, respectively. The maximum lower bound in each case ($\gamma_a$, $\gamma_b$, and $\gamma_c$) is represented by the black dashed line. It can be seen that $\Delta \gamma(H_1) \triangleq \gamma_b - \gamma_a < \gamma_c - \gamma_a \triangleq \Delta \gamma(H_2)$. The potential in each case ($P_a$, $P_b$, and $P_c$) is represented by the sum of the gray parts of the intervals. It can be seen that $\Delta P(H_1) \triangleq P_a - P_b < P_a - P_c \triangleq \Delta P(H_2)$.}
\label{fig:BoundsHypotheses}
\vspace{-15pt}
\end{figure}

In order to estimate the potential reduction $\widehat{\Delta P}(H)$ expected when hypothesis $H$ is refined (as required by the heuristic), we need to first define a few quantities (Fig. \ref{fig:BoundsHypotheses}). We define the \emph{margin} $M_n(H)$ of a hypothesis $H$ after $n$ cycles have been spent on it, as the difference between its bounds, i.e., $M_n(H) \triangleq \overline{L}_n(H) - \underline{L}_n(H)$. Then we define the reduction of the margin of this hypothesis during its $n$-th refinement as ${\Delta M}_n(H) \triangleq M_{n-1}(H)-M_n(H)$. It can be seen that this quantity is positive, and because in general early refinements produce larger margin reductions than later refinements, it has a general tendency to decrease. Using this quantity we \emph{predict} the reduction of the margin in the next refinement using an exponentially weighted moving average, $\widehat{\Delta M}_{n+1}(H) \triangleq \alpha$ $ \widehat{\Delta M}_n(H)+ (1-\alpha) \Delta{M}_n(H)$, where $0<\alpha<1$. This moving average is initialized as ${\widehat{\Delta M}}_0(H)=\beta M_0(H)$. (In this work we used $\alpha =0.9$ and $\beta =0.25$.)

The predicted potential reduction $\Delta P(H)$ depends on whether the refinement of $H$ is expected to increase $\gamma $ or not: when $\gamma $ increases, every term in \eqref{eq:Potential} is reduced; when it does not, only the term corresponding to $H$ is reduced (compare figures \ref{fig:BoundsHypotheses}b and \ref{fig:BoundsHypotheses}c). Let us assume that the reduction in the upper bound, $\overline{L}_n(H) - \overline{L}_{n+1}(H)$, is equal to the increase in the lower bound, $\underline{L}_{n+1}(H) - \underline{L}_n(H)$, which is therefore predicted to be equal to $\widehat{\Delta M}_{n+1}(H) / {2}$. Let us define $\Delta \gamma (H)$ to be the increase in $\gamma$ when $H$ is refined, thus $\Delta \gamma (H) \triangleq \max\{\underline{L}_n(H) + \widehat{\Delta M}_{n+1}(H) / {2} - \gamma, 0\}$. Then the following expression is an estimate for the potential reduction when $H$ is refined,
\begin{equation}
\label{eq:DeltaPotential}
\widehat{\Delta P}(H)=
\left\{ \begin{array}{ll}
	\widehat{\Delta M}_{n+1}(H) / 2 + |\mathbb{A}|\Delta \gamma(H), & \ \text{if} \ \Delta \gamma (H)>0 \\ 
	\widehat{\Delta M}_{n+1}(H) / 2, & \ \text{otherwise}.
\end{array} \right.
\end{equation}
As mentioned before, the hypothesis that maximizes this quantity is the one selected to be refined next. 

The algorithm used by the FoAM is thus the following (a detailed explanation is provided immediately afterwards):

\begin{algorithmic}[1]
\STATE $\gamma \gets -\infty$
\FOR{$i = 1$ to $N_H$}
	\STATE $\left[\underline{L}(H_i),\overline{L}(H_i)\right]  \gets$ ComputeBounds$(H_i)$
	\IF {$\underline{L}(H_i) > \gamma$}  
		\STATE $\gamma \gets \underline{L}(H_i)$
	\ENDIF
	\IF {$\overline{L}(H_i) > \gamma$}  
		\STATE $\widehat{\Delta M}_0(H_i) \gets \beta M_0(H_i)$
		\STATE Compute $\widehat{\Delta P}(H_i)$
		\STATE $A$.Insert$(H_i, \widehat{\Delta P}(H_i))$
	\ENDIF
\ENDFOR

\WHILE{not reached Termination Conditions} 
	\STATE $H_i \gets A$.GetMax$()$
	\IF {$\overline{L}(H_i) > \gamma$}
		\STATE $\left[\underline{L}(H_i),\overline{L}(H_i)\right]  \gets$ RefineBounds$(H_i)$
		\STATE Compute $\widehat{\Delta P}(H_i)$
		\STATE $A$.Insert$(H_i, \widehat{\Delta P}(H_i))$
		\IF {$\underline{L}(H_i) > \gamma$}  
			\STATE $\gamma \gets \underline{L}(H_i)$
		\ENDIF
	\ENDIF
\ENDWHILE
\end{algorithmic}

The first stage of the algorithm is to initialize the bounds for all the hypotheses (line 3 above), use these bounds to estimate the expected potential reduction for each hypothesis (lines 8-9), and to insert the hypotheses in the priority queue $A$ using the potential reduction as the key (line 10). This priority queue $A$, supporting the usual operations \texttt{Insert} and \texttt{GetMax}, contains the hypotheses that are active at any given time. The \texttt{GetMax} operation, in particular, is used to efficiently find the hypothesis that, if refined, is expected to produce the greatest potential reduction. During this first stage the maximum lower bound $\gamma$ is also initialized (lines 1 and 4-6).

In the second stage of the algorithm, hypotheses are selected and refined alternately until termination conditions are rea\-ched. The next hypothesis to refine is simply obtained by extracting from the priority queue $A$ the hypothesis that is expected to produce, if refined, the greatest potential reduction (line 14). If this hypothesis is still viable (line 15), its bounds are refined (line 16), its expected potential reduction is recomputed (line 17), and it is reinserted into the queue (line 18). If necessary, the maximum lower bound is also updated (lines 19-21). One issue to note in this procedure is that the potential reductions used as key when hypotheses are inserted in the queue $A$ are outdated once $\gamma$ is modified. Nevertheless this approximation works well in practice and allows a complete hypothesis selection/refinement cycle (lines 13-23) to run in $O(\log |A|)$ where $|A|$ is the number of active hypotheses. This complexity is determined by the operations on the priority queue.

Rother and Sapiro \cite{RotICCV09} have suggested a different heuristic to select the next hypothesis to refine. Their heuristic consists on selecting the hypothesis whose current upper bound is greatest. This heuristic, however, in general required more cycles than the heuristic we are proposing here. To see why, consider the case in which, after some refinement cycles, two active hypotheses $H_1$ and $H_2$ still remain. Suppose that $H_1$ is better than $H_2$ ($\overline{L}_{n_1}(H_1)\gg \overline{L}_{n_2}(H_2)$), but at the same time it has been more refined ($n_1 \gg n_2$). As mentioned before, because of the decreasing nature of $\Delta M$, in these conditions we expect ${\Delta M}_{n_2}(H_2) \gg {\Delta M}_{n_1}(H_1)$. Therefore, if we chose to refine $H_1$ (as in \cite{RotICCV09}) many more cycles will be necessary to distinguish between the hypotheses than if we had chosen to refine $H_2$ (as in the heuristic explained above). However, the strategy of choosing the less promising hypothesis is only worthwhile when there are few hypotheses remaining, since computation in that case is invested in a hypothesis that is ultimately discarded. The desired behavior is simply and automatically obtained by minimizing the potential defined in \eqref{eq:Potential}, and this ensures that computation is spent sensibly.

\section{Definition of the Problem}
\label{sec:ProblemDefinition}

The FoAM described in the previous section is a general algorithm that can be used to solve many different problems, as long as: 1) the problems can be formulated as selecting the hypothesis that maximizes some evidence function within a set of hypotheses, and 2) a suitable BM can be defined to bound this evidence function. To illustrate the use of the FoAM to solve a concrete problem, in this section we define the problem, and in Section \ref{sec:BMDirectProblem} we derive a BM for this particular problem.

Given an input image $I:\Omega \to \mathbb{R}^c$ ($c \in \mathbb{N}, c > 0$) in which there is a single ``shape'' corrupted by noise, the problem is to estimate the class $K$ of the shape, its pose $T$, and recover a noiseless version of the shape. This problem arises, for example, in the context of optical character recognition \cite{Fuj08} and shape matching \cite{Vel01}. For clarity it is assumed in this section that $\Omega \subset \mathbb{Z}^2$ (i.e., the image is composed of discrete pixels arranged in a 2D grid).

To solve this problem using a H\&B algorithm, we define one hypothesis $H$ for \emph{every possible} pair $(K,T)$. By selecting a hypothesis, the algorithm is thus estimating the class $K$ and pose $T$ of the shape in the image. As we will later show, in the process a noiseless version of the shape will also be obtained.

In order to define the problem more formally, suppose that the image domain $\Omega$ contains $n$ pixels, $\vec{x}_1,\dots,$ $\vec{x}_n$, and that there are $N_K$ distinct possible shape classes, each one characterized by a known shape prior $B_K$ ($1 \le K \le N_K$) defined on the whole discrete plane $\mathbb{Z}^2$, also containing discrete pixels. Each shape prior $B_K$ specifies, for each pixel $\vec{x}' \in \mathbb{Z}^2$, the probability that the pixel belongs to the shape $q$, $p_{B_K}(\vec{x}') \triangleq P(q(\vec{x}')=1 | K)$, or to the complement of the shape, $P(q(\vec{x}')=0|K)=1-p_{B_K}(\vec{x}')$. We assume that $p_{B_K}$ is zero everywhere, except (possibly) in a region $\Omega_K \subset \mathbb{Z}^2$ called the \emph{support} of $p_{B_K}$. We will say that a pixel $\vec{x}'$ belongs to the \texttt{Foreground} if $q(\vec{x}')=1$, and to the \texttt{Background} if $q(\vec{x}')=0$ (\texttt{Foreground} and \texttt{Background} are the labels of the two possible \emph{states} of a shape for each pixel). 

Let $T \in \{T_1,\dots ,T_{N_T}\}$ be an affine transformation in $\mathbb{R}^2$, and call $B_H$ (recall that $H=(K,T)$) the shape prior that results from transforming $B_K$ by $T$, i.e., $p_{B_H}(\vec{x}) \triangleq P(q(\vec{x}) = 1| H) \triangleq p_{B_K}(T^{-1}\vec{x})$ (disregard for the moment the complications produced by the misalignment of pixels). The state $q(\vec{x})$ in a pixel $\vec{x}$ is thus assumed to depend only on the class $K$ and the transformation $T$ (in other words, it is assumed to be conditionally independent of the states in the other pixels, given the hypothesis $H$).

Now, suppose that the shape $q$ is not observed directly, but rather that it defines the distribution of a feature (e.g., colors, edges, or in general any feature) to be observed at a pixel. In other words, if a pixel $\vec{x}$ belongs to the background (i.e., if $q(\vec{x})=0$), its feature $f(\vec{x})$ is distributed according to the probability density function $p_{\vec{x}}(f(\vec{x}) | q(\vec{x})=0)$, while if it belongs to the foreground (i.e., if $q(\vec{x})=1$), $f(\vec{x})$ is distributed according to $p_{\vec{x}}(f(\vec{x}) | q(\vec{x})=1)$ (the subscript $\vec{x}$ in $p_{\vec{x}}$ was added to emphasize the fact that the probability of observing a feature $f(\vec{x})$ at a pixel $\vec{x}$ depends on the state of the pixel $q(\vec{x})$ \emph{and} on the particular pixel $\vec{x}$, or in other terms, $p_{\vec{x}}(f_0 | q_0) \ne p_{\vec{y}}(f_0 | q_0)$ if $\vec{x} \ne \vec{y}$ and $f_0$ and $q_0$ are two arbitrary values of $f$ and $q$, respectively). This feature $f(\vec{x})$ is assumed to be independent of the feature $f(\vec{y})$ and the state $q(\vec{y})$ in every other pixel $\vec{y}$, given $q(\vec{x})$.

\begin{figure}
\vspace{-5pt}
\begin{center}
\includegraphics[width=0.6\columnwidth, bb=0pt 0pt 166pt 188pt]{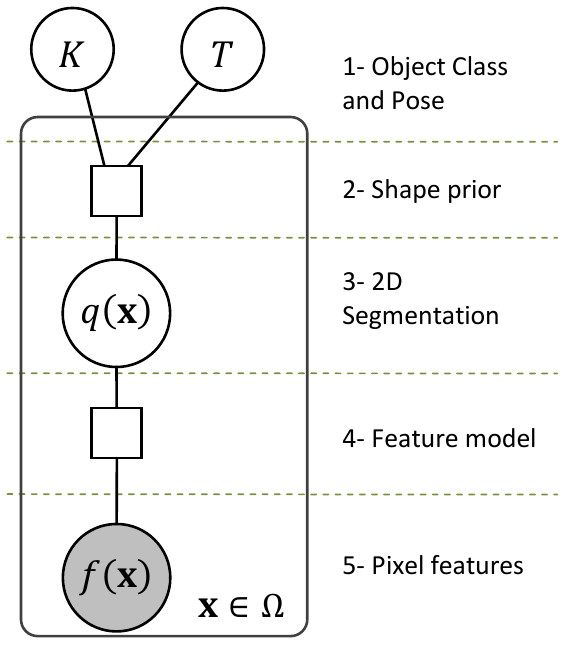}
\end{center}
\vspace{-10pt}
\caption{Factor graph proposed to solve our problem of interest. A \emph{factor graph}, \cite{Bis06}, has a \emph{variable node} (circle) for each variable, and a \emph{factor node} (square) for each factor in the system's joint probability. Factor nodes are connected to the variable nodes of the variables in the factor. Observed variables are shaded. A plate indicates that there is an instance of the nodes in the plate for each element in a set (indicated on the lower right). The plate in this graph hides the existing loops. See text for details.}
\vspace{-15pt}
\label{fig:FactorGraph}
\end{figure}

The conditional independence assumptions descri\-bed above can be summarized in the factor graph of Fig. \ref{fig:FactorGraph} (see \cite{Bis06} for more details on factor graphs). It then follows that the joint probability of all pixel features $f$, all states $q$, and the hypothesis $H=(K,T)$, is
\begin{equation}
p(f,q,H)=P(H) \prod_{\vec{x} \in \Omega} {P_{\vec{x}}(f(\vec{x}) | q(\vec{x})) P(q(\vec{x}) | H)}.
\label{eq:JointProbability}
\end{equation}

Then, our goal can be simply stated as solving
\begin{align}
\label{eq:GoalDirectProblem}
\mathop{\max }_{q,H} p(f,q,H) & = \mathop{\max }_{H \in \mathbb{H}} L''(H), \ \ \text{with} \\
L''(H) & \triangleq {\mathop{\max}_{q} p(f,q,H)}.
\label{eq:LPrimePrime}
\end{align}
We could solve this problem na\"{i}vely by computing \eqref{eq:LPrimePrime} for every $H \in \mathbb{H}$. However, to compute \eqref{eq:LPrimePrime} all the pixels in the image (or at least all the pixels in the support of $\Omega_K$) need to be processed in order to evaluate the product in \eqref{eq:JointProbability} (since the solution $q*$ that maximizes \eqref{eq:LPrimePrime} can be written explicitly). Because this might be very expensive, we need a BM to evaluate the evidence without having to process every pixel in the image. 

Therefore, instead of using this na\"{i}ve approach, we will use an H\&B algorithm to find the hypothesis $H$ that maximizes an expression simpler than $L''(H)$, that is equivalent to it (in the sense that it has the same maxima). This simpler expression is what we have called the evidence, $L(H)$, and will be derived from \eqref{eq:LPrimePrime} in Section \ref{sec:EvidenceDirectProblem}. Before deriving the evidence, however, in Section \ref{sec:TheoryOfShapes} we present the mathematical framework that will allow us to do that, and later to develop the BM for this evidence.

\section{A new theory of shapes}
\label{sec:TheoryOfShapes}

As mentioned before, H\&B algorithms have two parts: a FoAM and a BM. The FoAM was already introduced in Section \ref{sec:FoAM}. While the same FoAM can be used to solve many different problems, each BM is specific to a particular problem. Towards defining the BM for the problem described in the previous section (that will be done in Section \ref{sec:BMDirectProblem}), in this section we introduce a mathematical framework that will allow us to compute the bounds.

\begin{figure}
\vspace{-5pt}
\includegraphics[width=\columnwidth, bb= 0pt 0pt 318pt 91pt]{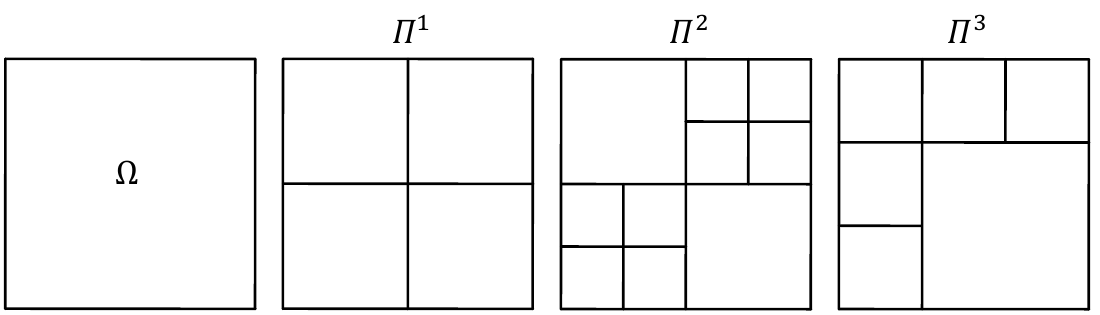}
\caption{The set $\Omega$ and three possible partitions of it. Each square represents a partition element. $\Pi^2$ is finer than $\Pi^1$ ($\Pi^2 \le \Pi^1$). $\Pi^3$ is not comparable with neither $\Pi^1$ nor with $\Pi^2$.}
\vspace{-15pt}
\label{fig:Partitions}
\end{figure}

To derive formulas to bound the evidence $L(H)$ of a hypothesis $H$ (the goal of the BM) for our specific problem of interest, we introduce in this section a framework to represent shapes and to compute bounds for their log-probability (in Section \ref{sec:EvidenceDirectProblem} we will show that this log-probability is closely related to the evidence). Three different shape representations will be introduced (Fig. \ref{fig:ShapeRepresentations}). Continuous shapes are the shapes that we would observe if our cameras (or 3D scanners) had ``infinite resolution.'' In that case it would be possible to compute the evidence $L(H)$ of a hypothesis with ``infinite precision" and therefore always select the (single) hypothesis whose evidence is maximum (except in concocted examples which have very low probability of occurring in practice). However, since real cameras and scanners have finite resolution, we introduce two other shape representations that are especially suited for this case: \emph{discrete} and \emph{semidiscrete} shape representations. Discrete shapes will allow us to compute a lower bound $\underline{L}(H)$ for the evidence of a hypothesis $H$. Semidiscrete shapes, on the other hand, will allow us to compute an upper bound $\overline{L}(H)$ for the evidence of a hypothesis $H$. Discrete and semidiscrete shapes are defined on partitions of the input image (i.e., non-overlapping regions that completely cover the image, see Fig. \ref{fig:Partitions}). Finer partitions result in tighter bounds, more computation, and the possibility of distinguishing more similar hypotheses. Coarser partitions on the other hand, result in looser bounds, less computation and more hypotheses that are indistinguishable (in the partition).

Previously we have assumed that the image domain, $\Omega \subset \mathbb{Z}^2$, consisted on discrete pixels arranged in a 2D grid. For reasons that will soon become clear, however, we assume from now on that the image domain $\Omega \subset \mathbb{R}^2$ is continuous. Thus, to discretize this continuouos domain we rely on ``partitions," defined next.

\begin{definition}[Partitions]
\label{def:Partitions}
Given a set $\Omega \subset {\mathbb R}^d$, a \emph{partition} $\Pi(\Omega)=\{{\Omega}_1,\dots,{\Omega}_n\}$ with ${\Omega}_i\ne \emptyset$, is a disjoint cover of the set $\Omega$ (Fig. \ref{fig:Partitions}). Formally, $\Pi(\Omega)$ satisfies
\begin{equation}
\bigcup^n_{i=1}{\Omega_i}=\Omega, \ \ \text{and} \  \ \Omega_i \cap \Omega_j=\emptyset \ \ \forall i\ne j. 
\label{eq:PartitionsConditions}
\end{equation}

A partition $\Pi(\Omega)=\{\Omega_1,\dots,\Omega_n\}$ is said to be \emph{uniform} if all the elements in the partition have the same measure $\left|\Omega_i\right|=\frac{\left|\Omega \right|}{n}$ for $i=1, \dots, n$. (Throughout this article we use the notation $\left|\Omega \right|$ to refer to, depending on the context, the measure or the cardinality of a set $\Omega$.) This measure is referred to as the \emph{unit size} of the partition. For $d=2$ and $d=3$, we will refer to the elements of the partition (${\Omega }_i$) as \emph{pixels} and \emph{voxels}, respectively.

Given two partitions $\Pi^1(\Omega)$ and $\Pi^2(\Omega)$ of a set $\Omega \subset {\mathbb R}^d$, $\Pi^2$ is said to be finer than $\Pi^1$, and $\Pi^1$ is said to be coarser than $\Pi^2$, if every element of $\Pi^2$ is a subset of some element of $\Pi^1$ (Fig. \ref{fig:Partitions}). We denote this relationship as $\Pi^2 \le \Pi^1$. Note that two partitions are not always comparable, thus the binary relationship ``$\le$'' defines a partial order in the space of all partitions.
\end{definition}

\subsection{Discrete shapes}
\label{sec:DiscreteShapes}
\begin{figure}
\includegraphics[width=\columnwidth, bb=0pt 0pt 1710pt 598pt]{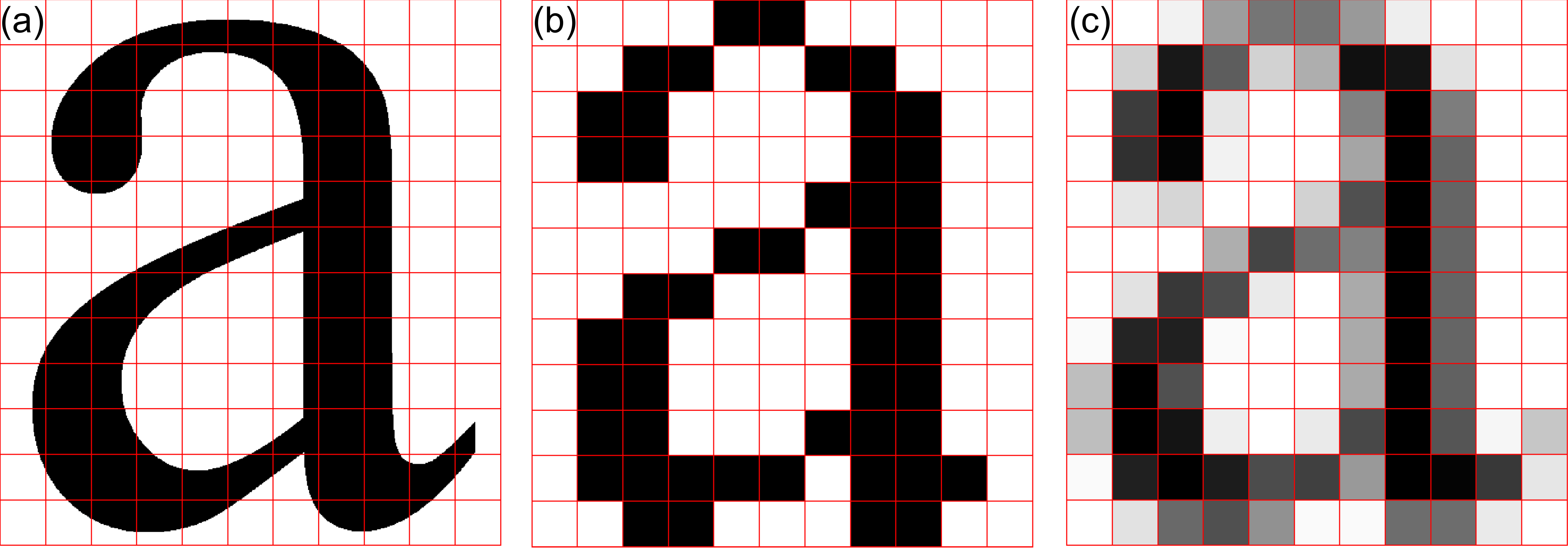}
\caption{The three different shape representations used in this article: the continuous shape $S$ (a), the discrete shape $\hat{S}$ (b), and the semidiscrete shape $\tilde{S}$ (c). $\hat{S}$ and $\tilde{S}$ are two approximations of $S$ in a finite partition (indicated by the red lines). The gray levels in $\tilde{S}$ represents how much of an element is full. See text for details.}
\vspace{-15pt}
\label{fig:ShapeRepresentations}
\end{figure}

\begin{definition}[Discrete shapes]
\label{def:DiscreteShape}
Given a partition $\Pi(\Omega)$ of the set $\Omega \subset {\mathbb R}^d$, the \emph{discrete shape} $\hat{S}$ (Fig. \ref{fig:ShapeRepresentations}b) is defined as the function $\hat{S}:\Pi(\Omega) \to \{0,1\}$.
\end{definition}

\begin{definition}[Log-probability of a discrete shape]
\label{def:LogPDiscreteShape}
Let $\hat{S}$ be a discrete shape in some partition $\Pi(\Omega)=\{\Omega_1,\dots,\Omega_n\}$, and let $\hat{B}=\{\hat{B}_1,\dots,\hat{B}_n\}$ be a family of independent Bernoulli random variables referred to as a discrete Bernoulli field (BF). Let $\hat{B}$ be characterized by the success rates $p_{\hat{B}}(i) \triangleq P({\hat{B}}_i=1) \in (\varepsilon,1-\varepsilon)$ for $i=1, \dots, n$, and $0<\varepsilon \ll 1$. To avoid the problems derived from assuming complete certainty, i.e. success rates of 0 or 1, following Cromwell's rule \cite{Lin85}, we will only consider success rates in the open interval $(\varepsilon,1-\varepsilon)$.

The \emph{log-probability of a discrete shape} is defined as
\begin{align}
\log {P(\hat{B}=\hat{S})} & \triangleq \sum_{i=1}^n{\log P\left(\hat{B}_i=\hat{S}(\Omega_i)\right)} \nonumber \\
& = \sum_{i=1}^n{\bigg[\left(1 - \hat{S}(\Omega_i)\right) \log{P\left(\hat{B}_i=0 \right)} + }\nonumber \\ 
& \qquad \qquad \hat{S}(\Omega_i) \log{P\left(\hat{B}_i=1 \right)} \bigg] \nonumber 
\end{align}
\begin{align}
\qquad \qquad = Z_{\hat{B}} + \sum_{i=1}^n{\hat{S}(\Omega_i)\delta_{\hat{B}}(i)},
\label{eq:LogPDiscreteShape}
\end{align}
where $Z_{\hat{B}} \triangleq \sum_{i=1}^n{\log \left(1-p_{\hat{B}}(i)\right)}$ is a constant and $\delta _{\hat{B}}(i) \triangleq {\log \left({p_{\hat{B}}(i)} / {\left(1-p_{\hat{B}}(i)\right)}\right)}$ is the logit function of $p_{\hat{B}}(i)$.
\end{definition}

The discrete BFs used in this work arise from two sources: background subtraction and shape priors. To compute a discrete BF ${\hat{B}}_f$ using the \emph{Background Subtraction} technique \cite{Mit04}, recall the probability densities $p_{\Omega_i}\left(f(\Omega_i) | q(\Omega_i)=0\right)$ and $p_{\Omega_i}\left(f(\Omega_i) | q(\Omega_i)=1\right)$ defined in Section \ref{sec:ProblemDefinition} to model the probability of observing a feature $f(\Omega_i)$ at a given pixel $\Omega_i$, depending on the pixel's state $q(\Omega_i)$. The success rates of the discrete BF $\hat{B}_f$ are thus defined as 
\begin{equation}
p_{\hat{B}_f}(i) \triangleq \frac{p_{\Omega_i}(f(\Omega_i) | q(\Omega_i) = 1)} {p_{\Omega_i}(f(\Omega_i) | q(\Omega_i)=0) + p_{\Omega_i}(f(\Omega_i) | q(\Omega_i)=1)}.
\label{eq:BFBackgrounSubtraction}
\end{equation}

To compute a discrete BF $\hat{B}_s$ associated with a \emph{discrete shape prior}, we can estimate the success rates $p_{\hat{B}_s}(i)$ of $\hat{B}_s$ from a collection of $N$ discrete shapes, $\hat{\Sigma}=\left\{\hat{S}_1,\dots,\hat{S}_N\right\}$, assumed to be aligned in the set $\Omega$. These discrete shapes can be acquired by different means, e.g., using a 2D or 3D scanner, for $d=2$ or $d=3$, respectively. The success rate $p_{\hat{B}_s}(i)$ of a particular Bernoulli variable $\hat{B}_i$ ($i=1, \dots, n$) is thus estimated from $\left\{\hat{S}_1(\Omega_i),\dots,\right.$ $\left. \hat{S}_N(\Omega_i)\right\}$ using the standard formula for the estimation of Bernoulli distributions \cite{Kay93},
\begin{equation}
p_{\hat{B}_s}(i)=\frac{1}{N}\sum^N_{j=1}{\hat{S}_j(\Omega_i)}.
\label{eq:BFFromTrainingSet}
\end{equation}

Discrete shapes, as in Definition \ref{def:DiscreteShape}, have two limitations that must be addressed to enable subsequent developments. First, the log-probability in \eqref{eq:LogPDiscreteShape} depends (implicitly) on the unit size of the partition (which is related to the image resolution), preventing the comparison of log-probabilities of images acquired at different resolutions (this will be further explained after Definition \ref{def:LogPContinuousShape}). Second, it was assumed in \eqref{eq:LogPDiscreteShape} that the Bernoulli variables $\hat{B}_i$ and the pixels $\Omega_i$ were perfectly aligned. However, this assumption might be violated if a transformation (e.g., a rotation) is applied to the shape. To overcome these limitations, and also to facilitate the proofs that will follow, we introduce next the second shape representation, that of a continuous shape.

\subsection{Continuous shapes}
\label{sec:ContinuousShapes}
\begin{definition}[Continuous shapes]
\label{def:ContinuousShape}
Given a set $\Omega \subset {\mathbb R}^d$, we define a \emph{continuous shape} $S$ to be a function $S:\Omega \to \{0,1\}$ (Fig. \ref{fig:ShapeRepresentations}a). We will often abuse notation and refer to the set $S=\left\{\vec{x} \in \Omega:S(\vec{x})=1\right\}$ also as the shape. To avoid pathological cases, we will require the set $S$ to satisfy two regularity conditions: 1) to be open (in the usual topology in ${\mathbb R}^d$ \cite{Kah95}) and 2) to have a boundary (as defined in \cite{Kah95}) of measure zero.
\end{definition}

Given a discrete shape $\hat{S}$ defined on a partition $\Pi(\Omega)=\left\{\Omega_1,\dots,\Omega_n\right\}$, the continuous shape $S(\vec{x}) \triangleq \hat{S}(\Omega_i) \ \forall \vec{x} \in \Omega_i$, is referred to as the continuous shape \emph{produced} by the discrete shape $\hat{S}$, and is denoted as $S \sim \hat{S}$ or $\hat{S} \sim S$. Intuitively, $S$ extends $\hat{S}$ from every element of $\Pi(\Omega)$ to every point $\vec{x} \in \Omega$.

We would like now to extend the definition of the log-probability of a discrete shape (in Definition \ref{def:LogPDiscreteShape}) to include continuous shapes. Toward this end we first introduce continuous BFs, which play in the continuous case the role that discrete BFs play in the discrete case.

\begin{definition}[Continuous Bernoulli Fields]
\label{def:ContinuousBF}
Given a set $\Omega \subset {\mathbb R}^d$, a \emph{continuous Bernoulli field} (or simply a BF) is the construction that associates a Bernoulli random variable $B_{\vec{x}}$ to every point $\vec{x} \in \Omega$. The \emph{success rate} for each variable in the field is given by the function $p_B(\vec{x}) \triangleq P(B_{\vec{x}}=1)$. The corresponding \emph{logit function} $\delta_B(\vec{x}) \triangleq {\log \left(\frac{p_B(\vec{x})} {1-p_B(\vec{x})}\right)}$ and \emph{constant term} $Z_B \triangleq \int_\Omega{{\log \left(1-p_B(\vec{x})\right) }d\vec{x}}$ are as in Definition \ref{def:LogPDiscreteShape}.
\end{definition}

We will only consider in this work functions $p_B(\vec{x})$ such that $\left|Z_B\right|<\infty$ and $\delta_B(\vec{x})$ is a measurable function \cite{Wil78}. Furthermore, since $\varepsilon<p_B(\vec{x})<1-\varepsilon \ \forall \vec{x} \in \Omega$, $\delta_B(\vec{x}) \in (-\delta_{max},\delta_{max}) \ \forall \vec{x} \in \Omega$, with $\delta_{max} \triangleq {\log  \left(\frac{1-\varepsilon }{\varepsilon}\right)}$. Note that a BF is \emph{not} associated to a \emph{continuous} probability density on $\Omega$ (e.g., it almost never holds that $\int_{\Omega }{p_B(\vec{x})\ d\vec{x}}=1$), but rather to a collection of \emph{discrete} probability distributions, one for each point in $\Omega$ (thus, it always holds that $P\left(B_{\vec{x}}=0\right)+P\left(B_{\vec{x}}=1\right)=1\ \forall \vec{x}\in \Omega$). 

Due to the finite resolution of cameras and scanners, continuous BFs cannot be directly obtained as discrete BFs were obtained in Definition \ref{def:LogPDiscreteShape}. In contrast, continuous BFs are obtained indirectly from discrete BFs (which are possibly obtained by one of the methods described in Definition \ref{def:LogPDiscreteShape}). Let $\hat{B}$ be a discrete BF defined on a partition $\Pi(\Omega)=\left\{\Omega_1,\dots,\Omega _n\right\}$. Then, for each partition element $\Omega_i$, and for each point $\vec{x}\in \Omega_i$, the success rate of the Bernoulli variable $B_{\vec{x}}$ is defined as $p_B(\vec{x}) \triangleq p_{\hat{B}}(i)$. The BF $B$ produced in this fashion will be referred to as the BF \emph{produced} by the discrete BF $\hat{B}$. Intuitively, $p_B$ extends $p_{\hat{B}}$ from every element of $\Pi(\Omega)$ to every point $\vec{x} \in \Omega$. Note that this definition is analogous to the definition of a continuous shape produced from a discrete shape in Definition \ref{def:ContinuousShape}.

Let $\Omega' \subset {\mathbb R}^d$ be a set referred to as the \emph{canonical set}, let $\Omega \subset {\mathbb R}^d$ be a second set referred to as the \emph{world set}, and let $T:\Omega' \to \Omega$ be a bijective transformation between these sets. Given a BF $B$ in $\Omega'$ with success rates $p_B(\vec{x})$, the transformed BF $B_T$ in $\Omega$ is defined as $B_T \triangleq B \circ T^{-1}$ with success rates $p_{B_T}(\vec{x}) \triangleq p_B\left(T^{-1}\vec{x}\right)$.

\begin{definition}[Log-probability of a continuous sha\-pe]
\label{def:LogPContinuousShape}
Let $B$ be a BF in $\Omega$ with success rates given by the function $p_B(\vec{x})$, let $S$ be a continuous shape also in $\Omega$, and let $u_o>0$ be a scalar called the \emph{equivalent unit size}. We define the \emph{log-probability that a shape $S$ is produced by a BF $B$}, by extension of the log-probability of discrete shapes in \eqref{eq:LogPDiscreteShape}, as
\begin{equation}
\log P(B=S) \triangleq \frac{1}{u_o} \left[Z_B + \int_{\Omega}{S(\vec{x}) \delta_B(\vec{x})\ d\vec{x}}\right],
\label{eq:LogPContinuousShape}
\end{equation}

\noindent where $Z_B$ and ${\delta }_B$ are respectively the constant term and the logit function of $B$.
\end{definition}

Several things are worth noting in this definition. First, note that if there is a uniform partition $\Pi(\Omega)=\left\{\Omega_1,\dots,\Omega_n\right\}$ with $\left|\Omega_i\right|=u_o\ \forall i$, and if the continuous shape $S$ and the BF $B$ are respectively produced by the discrete shape $\hat{S}$ and the discrete Bernoulli field $\hat{B}$ defined on $\Pi(\Omega)$, then $\log P(B=S)= \log P\left(\hat{B}=\hat{S}\right)$. For this reason we said that the definition in \eqref{eq:LogPContinuousShape} extends the definition in \eqref{eq:LogPDiscreteShape}. However, keep in mind that in the case of a continuous shape \eqref{eq:LogPContinuousShape} \emph{is not a log-probability in the traditional sense}, but rather it extends the definition to cases in which $S(\vec{x})$ is not produced from a discrete shape and $\delta_B(\vec{x})$ is not piecewise constant in a partition of $\Omega$.

Second, note that while \eqref{eq:LogPContinuousShape} provides the ``log-pro\-ba\-bi\-lity" density that a given continuous shape is produced by a BF, sampling from a BF is not guaranteed to produce a continuous shape (because the resulting set might not satisfy the regularity conditions in Definition \ref{def:ContinuousShape}). Nevertheless, this is not an obstacle since in this work we are only interested in computing log-probabilities of continuous shapes that are given, not on sampling from BFs.

Third, note in \eqref{eq:LogPContinuousShape} that the log-probability of a continuous shape is the product of two factors: 1) the inverse of the unit size, which only depends on the partition (but not on the shape); and 2) a term (in brackets) that does not depend on the partition. In the case of continuous shapes, $u_o$ in the first factor is not the unit size of the partition (there is no partition defined in this case) but rather a scalar defining the unit size of an \emph{equivalent partition} in which the range of log-probability values obtained would be comparable. The second factor is the sum of a constant term that only depends on the BF $B$, and a second term that also depends on the continuous shape $S$.

Fourth, the continuous shape representation in Definition \ref{def:ContinuousShape} overcomes the limitations of the discrete representation pointed out above. More specifically, by considering continuous shapes, \eqref{eq:LogPContinuousShape} can be computed even if a discrete shape and a discrete BF are defined on partitions that are not aligned, allowing us greater freedom in the choice on the transformations ($T$) that can be applied to the BF. Furthermore, the role of the partition is ``decoupled'' from the role of the BF and the shape, allowing us to compute \eqref{eq:LogPContinuousShape} independently of the resolution of the partitions. 

\subsection{Semi-discrete shapes}
\label{sec:SemiDiscreteShapes}
As mentioned at the beginning of this section, discrete shapes will be used to obtain a lower bound for the log-probability of a continuous shape. Unfortunately, upper bounds for the log-probability derived using discrete shapes are not very tight. For this reason, to obtain upper bounds for this log-probability, we need to introduce the third shape representation, that of semidiscrete shapes.

\begin{definition}[Semidiscrete shapes]
\label{def:SemidiscreteShape}
Given a partition $\Pi(\Omega)=\left\{\Omega_1,\dots,\Omega_n\right\}$ of the set $\Omega \subset {\mathbb R}^d$, the \emph{semidiscrete shape} $\tilde{S}$ is defined as the function $\tilde{S}:\Pi(\Omega) \to [0,|\Omega|]$, that associates to each element $\Omega_i$ in the partition a real number in the interval $[0,\left|\Omega _i\right|]$, i.e., $\tilde{S}(\Omega_i) \in [0,\left|\Omega_i\right|]$ (Fig. \ref{fig:ShapeRepresentations}c).
\end{definition}

Given a continuous shape $S$ in $\Omega$, we say that this shape \emph{produces} the semidiscrete shape $\tilde{S}$, denoted as $S \sim \tilde{S}$ or $\tilde{S} \sim S$, if $\tilde{S}(\Omega_i)=|S \cap \Omega_i|$ for $i=1, \dots, n$. An intuition that will be useful later to understand the derivation of the upper bounds is that a semidiscrete shape produced from a continuous shape ``remembers'' the measure of the continuous shape inside each element of the partition, but ``forgets" where exactly the shape is located inside the element.

Given two continuous shapes in $\Omega$, $S_1$ and $S_2$, that produce the the same semidiscrete shape $\tilde{S}$ in the partition $\Pi(\Omega)$, we say the these continuous shapes are related (denoted as $S_1 \sim S_2$). The nature of this relationship is explored in the next proposition.

\begin{proposition}[Equivalent classes of continuous shapes]
\label{prop:EquivalentClassesOfShapes}
The relationship ``$\sim$'' defined above is an equivalence relation.
\begin{proof}
\hspace{-4pt}: The proof of this proposition is trivial from Definition \ref{def:SemidiscreteShape}. \qed
\end{proof}
\end{proposition}
We will say that these continuous shapes are equivalent in the partition $\Pi(\Omega)$, and by extension, we will also say that they are equivalent to $\tilde{S}$ (i.e., $S_i \sim \tilde{S}$).

\begin{proposition}[Relationships between shape representations]
\label{prop:RelationshipsBetweenShapes}
Let $\Pi(\Omega)=\left\{\Omega_1,\dots,\Omega_n\right\}$ be an arbitrary partition of a set $\Omega$, and let $\widehat{\mathbb S}(\Pi)$ and $\widetilde{\mathbb S}(\Pi)$ be the sets of all discrete and semidiscrete shapes, respectively, defined on $\Pi(\Omega)$. Let ${\mathbb S}$ be the set of all continuous shapes in $\Omega $. Then,
\begin{align}
\label{eq:DiscreteShapesInContinuousShapes}
\left\{S:S \sim \hat{S}, \hat{S} \in \widehat{\mathbb S}(\Pi)\right\} & \subset \mathbb{S},\ \  \text{and}, \\
\label{eq:SemiDiscreteShapesEqualContinuousShapes}
\left\{S:S \sim \tilde{S},\tilde{S} \in \widetilde{\mathbb S}(\Pi)\right\} & = \mathbb{S}.
\end{align}

\begin{proof}
\hspace{-4pt}: The proof of this proposition is trivial from definitions \ref{def:DiscreteShape}, \ref{def:ContinuousShape}, and \ref{def:SemidiscreteShape}. \qed
\end{proof}
\end{proposition}

\subsection{LCDFs and summaries}
\label{sec:LCDFSandSummaries}
So far we have introduced three different shape representations and established relationships among them. In this section we introduce the concepts of logit cumulative distribution functions (LCDFs) and summaries. These concepts will be necessary to use discrete and semidiscrete shapes to bound the log-probability of continuous shapes, and hence, to bound the evidence.

Intuitively, a LCDF ``condenses'' the ``information'' of a BF in a partition element into a monotonous function. A summary then further ``condenses'' this ``information'' into a single vector of fixed length. Importantly, the summary of a BF in a partition element can be used to bound the evidence $L(H)$ and can be computed in constant time, regardless of the number of pixels in the element. 

After formally defining LCDFs and summaries below, we will prove in this Section some of the properties that will be needed in Section \ref{sec:BMDirectProblem} to compute lower and upper bounds for $L(H)$. In the remainder of this section, unless stated otherwise, all partitions, shapes and BFs are defined on a set $\Omega \subset {\mathbb R}^d$.

\begin{definition}[Logit cumulative distribution function]
\label{def:LCDFs}
Given the logit function $\delta_B$ of a BF $B$ and a partition $\Pi$, the \emph{logit cumulative distribution function} (or LCDF) of the BF $B$ in $\Pi$ is the collection of functions $D_B={\left\{D_{B,\omega}\right\}}_{\omega \in \Pi}$, where each function $D_{B,\omega}:[-\delta_{max},\delta_{max}] \to \left[0,\left|\omega\right|\right]$ is defined as
\begin{equation}
\label{eq:LCDF_D}
D_{B,\omega}(\delta) \triangleq \left|\left\{\vec{x} \in \omega:\delta_B(\vec{x})<\delta \right\}\right| \ (\omega \in \Pi).
\end{equation}
\end{definition}

It must be noted that this definition is consistent, since from Definition \ref{def:ContinuousBF}, the logit function is measurable. The LCDF is named by analogy to the probability cumulative distribution function, but must not be confused with it. Informally, a LCDF ``condenses'' the information of the BF by ``remembering'' the values taken by the logit function inside each partition element, but ``forgetting'' where those values are inside the element. This relationship between BFs and their LCDFs is analogous to the relationship between continuous and semidiscrete shapes.

Equation \eqref{eq:LCDF_D} defines a non-decreasing and possibly discontinuous function. To see that this function is non-decreasing, note that the set $\chi_1 \triangleq \{\vec{x} \in \omega:\delta_B(\vec{x})<\delta_1 \}$ is included in the set $\chi_2 \triangleq \{\vec{x} \in \omega:\delta_B(\vec{x})<\delta_2 \}$ if $\delta_1 \le \delta_2$. Moreover, this function is not necessarily strictly increasing because it is possible to have $D_{B,\omega}(\delta_1)=D_{B,\omega}(\delta_2)$ with $\delta_1<\delta_2$ if the measure of the set $\{\vec{x} \in \omega: \delta_1 \le \delta_B(\vec{x}) < \delta_2  \}$ is zero. To see that the function defined in \eqref{eq:LCDF_D} can be discontinuous, note that this function will have a discontinuity whenever the function $\delta_B(\vec{x})$ is constant on a set of measure greater than zero.

Later in this section we will use the inverse of a LCDF, $D^{-1}_{B,\omega}(a)$. If $D_{B,\omega}(\delta)$ were strictly increasing and continuous, we could simply define $D^{-1}_{B,\omega}(a)$ as the unique real number $\delta \in \left[-\delta_{max},\delta_{max}\right]$ such that $D_{B,\omega}(\delta)=a$. For general LCDFs, however, this definition does not produce a value for every $a \in \left[0,\left|\omega\right|\right]$. Instead we use the following definition.

\begin{figure}
\vspace{-5pt}
\includegraphics[width=\columnwidth, bb=0pt 0pt 304pt 181pt]{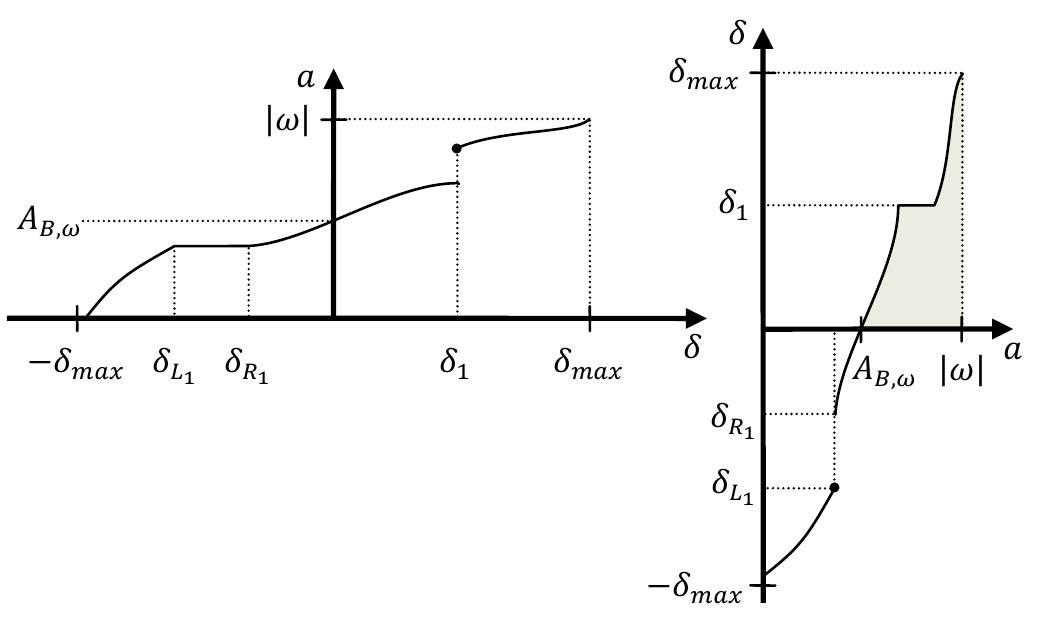}
\vspace{-15pt}
\caption{Plot of the LCDF $a=D_{B,\omega}(\delta)$ (left) and its inverse $\delta=D^{-1}_{B,\omega}(a)$ (right). The shaded area under the curve on the right is the maximal value of the integral on the rhs of \eqref{eq:LemaPart1} for any BF $B \sim D_{B,\omega}$ (see Lemma \ref{lem:PropertiesLCDFsAndSummaries}).}
\vspace{-15pt}
\label{fig:LCDFs}
\end{figure}

\begin{definition}[Inverse LCDF]
\label{def:InverseLCDF}
The \emph{inverse LCDF} $D^{-1}_{B,\omega}(a)$ is defined as (see Fig. \ref{fig:LCDFs})
\begin{equation}
\label{eq:InverseLCDF}
D^{-1}_{B,\omega}(a) \triangleq \inf \left\{\delta:D_{B,\omega}(\delta) \ge a\right\}.
\end{equation}
To avoid pathological cases, we will only consider in this work LCDFs whose inverse is continuous almost everywhere (note that this imposes an additional restriction on the logit function).
\end{definition}

\begin{definition}[X-axis crossing point]
\label{def:XCrossingPoint}
We define the \emph{x-axis crossing point}, $A_{B,\omega}$, as
\begin{equation}
\label{eq:XCrossingPoint}
A_{B,\omega} \triangleq \frac{D_{B,\omega}(0^-)+D_{B,\omega}(0^+)}{2}.
\end{equation}
\end{definition}
This quantity has the property that $D^{-1}_{B,\omega}(a) \ge 0$ if $a \ge A_{B,\omega}$, and $D^{-1}_{B,\omega}(a) \le 0$ if $a \le A_{B,\omega}$.

\begin{definition}[Summaries]
\label{def:Summaries}
Given a partition $\Pi$, a \emph{summary} of a LCDF $D_B={\left\{D_{B,\omega}\right\}}_{\omega \in \Pi}$ in the partition is a functional that assigns to each function $D_{B,\omega}$ ($\omega \in \Pi$) in the LCDF a vector $Y_{B,\omega} \in {\mathbb R}^{d_Y}$. The name ``summary'' is motivated by the fact that the ``infinite dimensional'' distribution is ``summarized'' by just $d_Y$ real numbers.

Two types of summaries are used in this article: \emph{m}-summaries and mean-summaries. Given $m>0$, an \emph{m-summary} assigns to each function $D_{B,\omega}$ in the LCDF the $(2m+1)$-dimensional vector $\tilde{Y}_{B,\omega} = \left[\tilde{Y}^{-m}_{B,\omega} \right. \dots$ ${\left. \tilde{Y}^m_{B,\omega}\right]}^T$ of equispaced samples of the LCDF, i.e.,
\begin{equation}
\label{eq:MSummary}
\tilde{Y}^j_{B,\omega} \triangleq D_{B,\omega}\left(\frac{j\delta_{max}}{m}\right) \ \ (j=-m,\dots ,m).
\end{equation}
Note that since the LCDF is known to be a non-decrea\-sing function, the information in the \emph{m}-summary can be used to bound the inverse LCDF (Fig. \ref{fig:BoundLCDFs}). Specifically, we know that $D^{-1}_{B,\omega}(a) \le \frac{\delta_{max}} {m} (j+1)$ for $a \in \left[\tilde{Y}^j_{B,\omega},\right.$ $\left. \tilde{Y}^{j+1}_{B,\omega}\right]$, which can be written as 
\begin{align}
D^{-1}_{B,\omega}(a) & \le \overline{D^{-1}_{Y_{B,\omega}}}(a) \nonumber \\
& \triangleq \frac{\delta_{max}}{m} J(a),\ \forall a \in \left[0,\left|\omega\right|\right],
\label{eq:DInverseUpper}
\end{align}
with
\begin{align}
J(a) & \triangleq \min \left\{j:\tilde{Y}^j_{B,\omega} \ge a\right\} \nonumber \\
& =\left|\left\{j:\tilde{Y}^j_{B,\omega}<a\right\}\right|-m.
\label{eq:J}
\end{align}
These bounds will be used in turn to compute an upper bound for $L(H)$.

\begin{figure}
\vspace{-10pt}
\begin{center}
\includegraphics[width=0.85\columnwidth, bb=0pt 0pt 217pt 145pt]{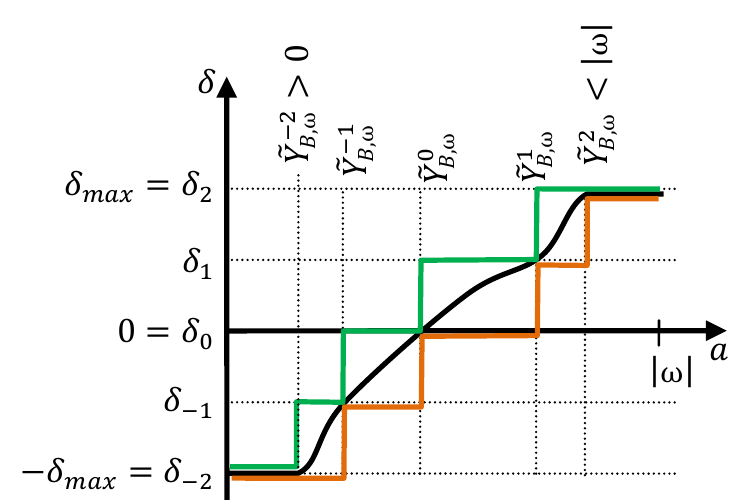}
\end{center}
\vspace{-10pt}
\caption{Lower (in orange) and upper (in green) bounds for the inverse LCDF $\delta=D_{B,\omega}^{-1}(a)$ (in black) computed from the \emph{m}-summaries with $m=2$. The $j$-th component of the summary, $\tilde{Y}_{B,\omega}^j$, is represented by a dotted vertical line at the horizontal position $a=\tilde{Y}_{B,\omega}^j$ (or where $D_{B,\omega}^{-1}(a)=\delta_j \triangleq j \delta_{max} / m$).}
\vspace{-15pt}
\label{fig:BoundLCDFs}
\end{figure}

The second type of summaries, referred to as mean-summaries, will be used to compute a lower bound of $L(H)$. The \emph{mean-summary} assigns to each function $D_{B,\omega}$ the scalar
\begin{equation}
\label{eq:MeanSummary}
\hat{Y}_{B,\omega} \triangleq \int_0^{|\omega|} {D^{-1}_{B,\omega}(a)\ da} = \int_{\omega}{\delta_B(\vec{x})\ d\vec{x}}.
\end{equation}
The last equality follows by setting $\tilde{S}(\omega)=|\omega|$ in \eqref{eq:LemaPart1} and is proved later in Lemma \ref{lem:PropertiesLCDFsAndSummaries}. Note that this equality provides the means to compute the mean-summary directly from the logit function, without having to compute the LCDF.
\end{definition}

One of the main properties of summaries is that, for certain kinds of sets, they can be computed in constant time regardless of the number of elements (e.g., pixels) in these sets. Next we show how to compute the \emph{mean}-summary $\hat{Y}_{B,\Phi}$ and the \emph{m}-summaries $\tilde{Y}_{B,\Phi}$ of a BF $B$ for the set $\Phi \subset \Omega$ (defined below).  For simplicity we assume that $\Omega \subset \mathbb{R}^2$, but the results presented here immediately generalize to higher dimensions. We also assume that $\Pi(\Omega)=\{\Omega_{1,1}, \Omega_{1,2}, \dots, …,\Omega_{n,n}\}$ is a uniform partition of $\Omega$ organized in rows and columns, where each partition element $\Omega_{i,j}$ (in the $i$-th row and the $j$-th column) is a square of area $|\Omega_{i,j}|=u_0$. We assume that $B$, defined by its logit function $\delta_B(\vec{x})$ ($\vec{x} \in \Omega)$, was obtained from a discrete shape prior $\hat{B}$ in $\Pi(\Omega)$ (as described in Definition \ref{def:ContinuousBF}), and therefore $\delta_B(\vec{x}) \triangleq \delta_{\hat{B}}(i,j) \ \forall \vec{x} \in \Omega_{i,j}$. And finally, we assume that $\Phi$ is an axis-aligned rectangular region containing only whole pixels (i.e., not parts of pixels). That is, 
\begin{equation}
\label{eq:Box}
\Phi \triangleq \bigcup_{
\begin{array}{c}
\scriptstyle{i_L \le i \le i_U} \\
\scriptstyle{j_L \le j \le j_U}
\end{array}}{\Omega_{i,j}}.
\end{equation}

In order to compute the \emph{mean}-summary $\hat{Y}_{B,\Phi}$, note that from \eqref{eq:MeanSummary},
\begin{align}
\hat{Y}_{B,\Phi} & = 
\sum_{\begin{array}{c}
\scriptstyle{i_L \le i \le i_U} \\
\scriptstyle{j_L \le j \le j_U}
\end{array}}{\int_{\Omega_{i,j}}{\delta_B(\vec{x})\ d\vec{x}}} = \nonumber \\
\label{eq:ToComputeMeanSummaries}
& = u_o \sum_{\begin{array}{c}
\scriptstyle{i_L \le i \le i_U} \\
\scriptstyle{j_L \le j \le j_U}
\end{array}}{\delta_{\hat{B}}(i,j)}.
\end{align}
The sum on the rhs of \eqref{eq:ToComputeMeanSummaries} can be computed \emph{in constant time} by relying on \emph{integral images} \cite{Vio01}, an image representation precisely proposed to compute sums in rectangular domains in constant time. To accomplish this, integral images precompute a matrix where each pixel stores the cumulative sum of the values in pixels with lower indices. The sum in \eqref{eq:ToComputeMeanSummaries} is then computed as the sum of four of these precomputed cumulative sums.

The formula to compute the \emph{m}-summary $\tilde{Y}_{B,\Phi}$ is similarly derived. From \eqref{eq:MSummary}, and since $\delta_B$ is constant inside each partition element, it holds for $k=-m, \dots,m$ that
\begin{align}
\tilde{Y}_{B,\Phi}^k & = \left | \left \{ \vec{x} \in \Phi : \delta_B(\vec{x}) < \frac{k\delta_{max}}{m} \right \} \right | =  u_o \left | \left \{ (i,j): \vphantom{\frac{j\delta_{max}}{m}} \right. \right. \nonumber \\
\label{eq:ToComputeMSummaries}
&  \left. \left.  i_L \le i \le i_U, j_L \le j \le j_U, \delta_{\hat{B}}(i,j) < \frac{k\delta_{max}}{m} \right \} \right | 
\end{align}
Let us now define the matrices $I_k$ ($k=-m, \dots, m$) as
\begin{equation}
I_k(i,j) \triangleq \left \{
\begin{array}{ll}
1, & \ \text{if} \ \delta_{\hat{B}}(i,j) < k \delta_{max} / m \\
0, & \ \text{otherwise}.
\end{array}
\right.
\end{equation}
Using this definition, \eqref{eq:ToComputeMSummaries} can be rewritten as
\begin{equation}
\tilde{Y}_{B,\Phi}^k = u_o \sum_{
\begin{array}{c}
\scriptstyle{i_L \le i \le i_U} \\
\scriptstyle{j_L \le j \le j_U}
\end{array}} {I_k(i,j)},
\end{equation}
which as before can be computed in $O(1)$ using integral images.

Before deriving formulas to bound $L(H)$, we need the following two results.

\begin{proposition}[Equivalent classes of BFs]
\label{prop:EquivalentClassesOfBFs}
Let $\Pi$ be a partition, let $B_1$ and $B_2$ be two BFs, and consider the following binary relations: 1) $B_1$ and $B_2$ are related (denoted as $B_1\sim_D B_2$) if they produce the same LCDF $D={\left\{D_{\omega}\right\}}_{\omega \in \Pi}$; and 2) $B_1$ and $B_2$ are related (denoted as $B_1\sim_Y B_2$) if they produce the same summary $Y={\left\{Y_{\omega}\right\}}_{\omega \in \Pi}$. Then, ``$\sim_D$" and ``$\sim_Y$" are equivalence relations.
\begin{proof}
\hspace{-4pt}: Immediate from definitions \ref{def:LCDFs} and \ref{def:Summaries}.\qed
\end{proof}
\end{proposition}

In the first case ($\sim_D$) we will say that these BFs are \emph{equivalent} in the partition $\Pi$ \emph{with respect to distributions}. Abusing the notation, we will also say that they are equivalent to (or \emph{compatible with}) the LCDF (i.e., $B_i\sim D$). Similarly, in the second case ($\sim_Y$) we will say that these BFs are \emph{equivalent} in the partition $\Pi$ \emph{with respect to summaries}. Abusing the notation, we will also say that they are equivalent to (or \emph{compatible with}) the summary (i.e., $B_i\sim Y$). Note that if two BFs are equivalent with respect to a LCDF, they are also equivalent with respect to any summary of the LCDF. The reverse, however, is not necessarily true.

\begin{lemma}[Properties of LCDFs and \emph{m}-Su\-mma\-ries]
\label{lem:PropertiesLCDFsAndSummaries}
Let $\Pi$ be an arbitrary partition, and let $\tilde{S}$ and $D={\left\{D_{\omega}\right\}}_{\omega \in \Pi}$ be respectively a semidiscrete shape and a LCDF in this partition. Then:

1. Given a BF $B$ such that $B\sim D$, it holds that (Fig. \ref{fig:LCDFs})
\begin{equation}
\label{eq:LemaPart1}
\mathop{\sup}_{S\sim \tilde{S}} \int_{\Omega}{\delta_B(\vec{x})S(\vec{x})\ d\vec{x}} = 
\sum_{\omega \in \Pi}{\int^{\left|\omega\right|}_{\left|\omega\right|-\tilde{S}(\omega)} {D^{-1}_{\omega}(a)\ da}}.
\end{equation}

2. Similarly, given a continuous shape $S$, such that $S\sim \tilde{S}$, it holds that
\begin{equation}
\label{eq:LemaPart2}
\mathop{\sup}_{B\sim D} \int_{\Omega}{\delta_B(\vec{x})S(\vec{x})\ d\vec{x}} =
\sum_{\omega \in \Pi} {\int^{\left|\omega\right|}_{\left|\omega\right|-\tilde{S}(\omega)} {D^{-1}_{\omega}(a)\ da}}.
\end{equation}

3. Moreover, for any $\alpha \in \left[0, |\omega|\right]$, the integrals on the rhs of \eqref{eq:LemaPart1} and \eqref{eq:LemaPart2} can be bounded as
\begin{align}
\int^{\left|\omega\right|}_{\alpha}{D^{-1}_{\omega}(a)\ da} \le \frac{\delta_{max}}{m} \left[ 
\vphantom{\sum^{m-1}_{j=J(\alpha)}{\left(\tilde{Y}^{j+1}_{\omega}\right)(j+1)}}
J(\alpha)\left(\tilde{Y}^{J(\alpha)}_{\omega}-\alpha \right) + \right. \nonumber \\
\label{eq:LemaPart3}
\left. \sum^{m-1}_{j=J(\alpha)}{\left(\tilde{Y}^{j+1}_{\omega}-\tilde{Y}^j_{\omega}\right)(j+1)} \right],
\end{align}
where $J(\alpha)$ is as defined in \eqref{eq:J}.

4. The rhs of \eqref{eq:LemaPart1} and \eqref{eq:LemaPart2} are maximum when $\tilde{S}(\omega)=|\omega|-A_{\omega}$, thus
\begin{equation}
\label{eq:LemaPart4}
\mathop{\sup}_{\tilde{S}} \sum_{\omega \in \Pi} {\int^{|\omega|}_{|\omega|-\tilde{S}(\omega)}{D^{-1}_{\omega}(a)\ da}} = 
\sum_{\omega \in \Pi} {\int^{|\omega|}_{A_{\omega}}{D^{-1}_{\omega}(a)\ da}}.
\end{equation}
\begin{proof}
\hspace{-4pt}: Due to space limitations this proof was included in the supplementary material. \qed
\end{proof}
\end{lemma}

This concludes the presentation of the general mathematical framework to compute and bound probabilities of shapes. This framework is used in the next section to bound the evidence for the problem defined in Section \ref{sec:ProblemDefinition}, and also in \cite{Rot102} to bound the evidence for a more complex problem.

\section{Bounding mechanism}
\label{sec:BMDirectProblem}

We are now ready to develop the BM for the problem defined in Section \ref{sec:ProblemDefinition}. This BM is the second part of the H\&B algorithm proposed to solve this problem. To develop the BM we proceed in three steps: first we derive an expression for the evidence $L(H)$ that the FoAM will maximize (Section \ref{sec:EvidenceDirectProblem}); second we show how to compute lower and upper bounds for the evidence for a given partition (Section \ref{sec:ComputationOfBounds}); and third, we show how to construct, incrementally, a sequence of increasingly finer partitions that will result in increasingly tighter bounds (Section \ref{sec:IncrementalBounds}). 

\subsection{Definition of the Evidence $L(H)$}
\label{sec:EvidenceDirectProblem}
In order to define the evidence $L(H)$ for the problem, we re-label the pixels of the image in \eqref{eq:JointProbability} (denoted previously by $\vec{x}$) as the elements of a uniform partition $\Pi$, and separate the factors according to its state $q(\omega)$. Hence \eqref{eq:JointProbability} is equal to
\begin{align}
P(H) \prod_{\omega \in \Pi} {} & P_{\omega}(f(\omega) | q(\omega)) P(q(\omega) | H) = \nonumber \\
P(H) \prod_{\omega \in \Pi} {} & 
{\big[P_{\omega}(f(\omega) | q(\omega) = 0) P(q(\omega) = 0 | H)\big]}^{1-q(\omega)} \times \nonumber \\
& \times {\big[P_{\omega}(f(\omega) | q(\omega) = 1) P(q(\omega) = 1 | H)\big]}^{q(\omega)}.
\label{eq:L0}
\end{align}
Defining a BF $B_f$ based on the features observed in the input image, with success rates given by
\begin{equation}
p_{B_f}(\omega) \triangleq \frac{P_{\omega}\left(f(\omega) | q(\omega) = 1 \right)} {P_{\omega}\left(f(\omega)|q(\omega)=0 \right) + P_{\omega}\left(f(\omega|q(\omega)=1 \right)},
\label{eq:BernoulliFieldInput}
\end{equation}
assuming that all hypotheses are equally likely, and dividing \eqref{eq:L0} by the constant (and known) term 
\begin{equation}
P(H) \prod_{\omega \in \Pi} {\left(P_{\omega}(f(\omega)|q(\omega)=0) + P_{\omega}(f(\omega)|q(\omega)=1) \right)},
\end{equation}
we obtain an expression equivalent to \eqref{eq:JointProbability},
\begin{align}
\prod_{\omega \in \Pi} {} & 
{\bigg[\big(1 - p_{B_f}(\omega)\big) \big(1 - p_{B_H}(\omega)\big) \bigg]}^{1-q(\omega)} \times \nonumber \\
& \times {\bigg[p_{B_f}(\omega) p_{B_H}(\omega) \bigg]}^{q(\omega)}
\label{eq:PreLPrime}
\end{align}
(recall from Section \ref{sec:ProblemDefinition} that $B_H$ is a BF with success rates $p_{B_H} \triangleq P(q(\omega) = 1 | H)$). This expression can be further simplified by taking logarithms and using the variables introduced in Definition \ref{def:LogPDiscreteShape} to yield
\begin{equation}
Z_{B_f} + Z_{B_H} + \sum_{\omega \in \Pi} {q(\omega) \big( \delta_{B_f}(\omega) + \delta_{B_H}(\omega) \big)}.
\label{eq:L1}
\end{equation}

Using the extension to the continuous domain explained in Definition \ref{def:LogPContinuousShape}, and substituting  \eqref{eq:L1} into \eqref{eq:LPrimePrime}, \eqref{eq:LPrimePrime} can be rewritten as
\begin{align}
L'(H) \triangleq \mathop{\sup}_{q} \frac{1}{u_o} \bigg[& Z_{B_f} + Z_{B_H} + \nonumber \\
& \int_{\Omega}{q(\vec{x}) \big( \delta_{B_f}(\vec{x}) + \delta_{B_H}(\vec{x}) \big) \ d\vec{x}} \bigg].
\label{eq:LPrime}
\end{align}
Now, since $Z_{B_f}$ and $u_o$ are constant for every hypothesis and every continuous shape $q$, maximizing this expression is equivalent to maximizing
\begin{equation}
\label{eq:L_H}
L(H) \triangleq Z_{B_H} + \mathop{\sup}_{q} \int_{\Omega}{q(\vec{x})\left(\delta_{B_f}(\vec{x})+\delta_{B_H}(\vec{x})\right)\ d\vec{x}}.
\end{equation}

This is the final expression for the evidence. Due to the very large number of pixels in a typical image produced by a modern camera, computing $L(H)$ directly as the integral in \eqref{eq:L_H} would be prohibitively expensive. For this reason, the next step is to derive bounds for \eqref{eq:L_H} that are cheaper to compute than \eqref{eq:L_H} itself, and are sufficient to discard most hypotheses. These bounds are derived in the next section.

\subsection{Derivation of the bounds}
\label{sec:ComputationOfBounds}
In the following two theorems we derive bounds for the evidence of a hypothesis that can be computed \emph{from summaries of the BFs} $B_f$ and $B_H$ (in \eqref{eq:L_H}), instead of computing them from the BFs directly. Because summaries can be computed in $O(1)$ for each element in a partition, bounds for a given partition can be computed in $O(n)$ (where $n$ is the number of elements in the partition), regardless of the actual number of pixels in the image.

\begin{theorem}[Lower bound for $L(H)$]
\label{theo:LowerBoundDirect}
Let $\Pi$ be a partition, and let $\hat{Y}_f={\left\{\hat{Y}_{f,\omega}\right\}}_{\omega \in \Pi}$ and $\hat{Y}_H={\left\{\hat{Y}_{H,\omega}\right\}}_{\omega \in \Pi}$ be the \emph{mean}-summaries of two unknown BFs in $\Pi$. Then, for any $B_f \sim \hat{Y}_f$ and any $B_H \sim \hat{Y}_H$, it holds that $L(H) \ge \underline{L}_{\Pi}(H)$, where
\begin{align}
\label{eq:LowerBound}
\underline{L}_{\Pi}(H) & \triangleq Z_{B_H} + \sum_{\omega \in \Pi}{\underline{\mathcal L}_{\omega}(H)}, \\
\label{eq:LowerBoundLocal}
\underline{\mathcal L}_{\omega}(H) & \triangleq \left(\hat{Y}_{f,\omega}+\hat{Y}_{H,\omega}\right)\hat{q}_*(\omega),
\end{align}
and $\hat{q}_*$ is a discrete shape in $\Pi$ defined as
\begin{equation}
\label{eq:DiscreteShapeLower}
\hat{q}_*(\omega) \triangleq \left\{ \begin{array}{ll}
	1, & \ \text{if} \left(\hat{Y}_{f,\omega} + \hat{Y}_{H,\omega}\right)>0 \\ 
	0, & \ \text{otherwise}. \end{array}
\right.
\end{equation}
\begin{proof}
\hspace{-4pt}: Due to space limitations this proof was included in the supplementary material. \qed
\end{proof}
\end{theorem}

\begin{theorem}[Upper bound for $L(H)$]
\label{theo:UpperBoundDirect}
Let $\Pi$ be a partition, and let $\tilde{Y}_f={\left\{\tilde{Y}_{f,\omega}\right\}}_{\omega \in \Pi}$ and $\tilde{Y}_H={\left\{\tilde{Y}_{H,\omega}\right\}}_{\omega \in \Pi}$ be the \emph{m}-summaries of two unknown BFs in $\Pi$. Let $\tilde{Y}_{f \bigoplus H,\omega}$ ($\omega \in \Pi$) be a vector of length $4m+2$ obtained by sorting the values in $\tilde{Y}_{f,\omega}$ and $\tilde{Y}_{H,\omega}$ (in ascending order), keeping repeated values, i.e.,
\begin{align}
\tilde{Y}_{f \bigoplus H,\omega} & \triangleq \left[\tilde{Y}^1_{f \bigoplus H,\omega},\dots,\tilde{Y}^{4m+2}_{f \bigoplus H,\omega} \right] \nonumber \\
\label{eq:SortedSummaries}
& \triangleq SortAscending\left(\tilde{Y}_{f,\omega} \cup \tilde{Y}_{H,\omega}\right).
\end{align}

Then, for any $B_f \sim \tilde{Y}_f$ and any $B_H \sim \tilde{Y}_H$, it holds that $L(H) \le \overline{L}_{\Pi}(H)$, where 
\begin{align}
\label{eq:UpperBound}
\overline{L}_{\Pi}(H) & \triangleq Z_{B_H} + \sum_{\omega \in \Pi} {\overline{\mathcal L}_{\omega}(H)}, \ \text{and}\\
\label{eq:UpperBoundLocal}
\overline{\mathcal L}_{\omega}(H) & \triangleq \frac{\delta_{max}}{m} \sum_{j=2m+1}^{4m+1}{(j-2m) \left( \tilde{Y}^{j+1}_{f \bigoplus H,\omega} - \tilde{Y}^j_{f \bigoplus H,\omega} \right)}.
\end{align}
It also follows that the continuous shape that maximizes \eqref{eq:L_H} is equivalent to a semidiscrete shape $\tilde{q}_*$ in $\Pi$ that satisfies
\begin{equation}
\label{eq:SemiDiscreteShapeUpper}
\tilde{q}_*(\omega) \in \left[|\omega| - \tilde{Y}^{2m+1}_{f \bigoplus H,\omega},|\omega| - \tilde{Y}^{2m}_{f \bigoplus H,\omega}\right] \ \forall \omega \in \Pi.
\end{equation}

\begin{proof}
\hspace{-4pt}: Due to space limitations this proof was included in the supplementary material. \qed
\end{proof}
\end{theorem}

Theorems 1 and 2 presented formulas to compute lower and upper bounds for $L(H)$, respectively, for a given partition $\Pi$. Importantly, these theorems also include formulas to compute a discrete shape $\hat{q}_*$ and a semidiscrete shape $\tilde{q}_*$ that approximate (in the partition $\Pi$) the continuous shape $q$ that solves \eqref{eq:L_H}. In the next section we show how to reuse the computation spent to compute the bounds for a partition ${\Pi}_k$, to compute the bounds for a finer partition ${\Pi}_{k+1}$.

\subsection{Incremental refinement of bounds}
\label{sec:IncrementalBounds}
Given a partition ${\Pi}_k$ containing $h_k$ elements, it can be seen in \eqref{eq:LowerBound} and \eqref{eq:UpperBound} that the bounds for the evidence corresponding to this partition can be computed in $O(h_k)$. In Section \ref{sec:FoAM}, however, we requested that the BM be able to compute these bounds in $O(1)$. In order to compute a sequence of progressively tigh\-ter bounds for a hypothesis $H$, where each bound is computed in $O(1)$, we inductively construct a sequence of progressively finer partitions of $\Omega$ for the hypothesis.

Let us denote by ${\Pi}_k(H) \triangleq \left\{\Omega_{H,1},\dots,\Omega_{H,h_k}\right\}$ the $k$-th partition in the sequence corresponding to $H$. Each sequence is defined inductively by
\begin{align}
{\Pi}_1(H) & \triangleq \{\Omega\},\ and \\
\label{eq:InductionPartition}
{\Pi}_{k+1}(H) & \triangleq \left[\Pi_k(H) \setminus \omega_k\right] \cup \pi(\omega_k), \ (k>0),
\end{align}
where $\omega_k \in \Pi_k(H)$ and $\pi(\omega_k)$ is a partition of $\omega_k$. For each partition $\Pi_k(H)$ in the sequence, lower ($\underline{L}_k(H)$) and upper ($\overline{L}_k(H)$) bounds for the evidence $L(H)$ could be computed in $O(k)$ using \eqref{eq:LowerBound} and \eqref{eq:UpperBound}, respectively. However, these bounds can be computed more efficiently by exploiting the form of \eqref{eq:InductionPartition}, as
\begin{equation}
\label{eq:LowerBoundInductive}
\underline{L}_{k+1}(H) = \underline{L}_k(H) - \underline{\mathcal L}_{\omega_k}(H) + \sum_{\omega \in \pi(\omega_k)} {\underline{\mathcal L}_{\omega}(H)}.
\end{equation}
(A similar expression for the upper bound $\overline{L}_{k+1}(H)$ can be derived.) If the partition of $\omega_k$, $\pi(\omega_k)$, is chosen to always contain a fixed number of sets (e.g., $\left|\pi (\omega_k)\right|=4 \ \forall k > 1$), then it can be seen in \eqref{eq:LowerBoundInductive} that $O(1)$ evaluations of \eqref{eq:LowerBoundLocal} are required to compute $\underline{L}_{k+1}(H)$.

While any choice of $\omega_k$ from $\Pi_k(H)$ in \eqref{eq:InductionPartition} would result in a new partition $\Pi_{k+1}(H)$ that is finer than $\Pi_k(H)$, it is natural to choose $\omega_k$ to be the set in $\Pi_k(H)$ with the greatest local margin ($\overline{\mathcal L}_{\omega_k}(H) - \underline{\mathcal L}_{\omega_k}(H)$) since this is the set responsible for the largest contribution to the total margin of the hypothesis ($\overline{L}_k(H) - \underline{L}_k(H)$). In order to efficiently find the set $\omega_k$ with the greatest local margin, we store the elements of a partition in a priority queue, using their local margin as the priority. Hence to compute $\Pi_{k+1}(H)$ from $\Pi_k(H)$ (in \eqref{eq:InductionPartition}) we need to \emph{extract} the element $\omega_k$ of the queue with the largest local margin, and then \emph{insert} each element in $\pi (\omega_k)$ into the queue. Taken together these steps have, depending on the implementation of the queue, complexity of at least $O(\log h_k)$ (where $h_k$ is the number of elements in the partition) \cite{Ron97}. In our case, however, it is not essential to process the elements in \emph{strictly} descending margin order. Any element with a margin close enough to the maximum margin would produce similar results. Moreover, in our case we know that the margins belong to the interval $(0,\overline{\mathcal L}_{\Omega }(H) - \underline{\mathcal L}_{\Omega }(H)]$ and that they tend to decrease with time. 

Based on these considerations we propose a queue implementation based on the \emph{untidy priority queue} of Yatziv et al. \cite{Yat06}, in which the operations \texttt{GetMax} and \texttt{Insert} both have complexity $O(1)$. This implementation consists of an array of buckets (i.e., singly-linked lists), where each bucket contains the elements whose margin is in an interval $I_j$. Specifically, suppose that the minimum and maximum margin of any element are known to be $\underline{\mathcal M}$ and $\overline{\mathcal M}$, respectively, and that $\rho >1$ is a constant (we chose $\rho =1.2$). The intervals are then defined to be $I_j\triangleq [\underline{\mathcal M}\rho^{j-1},\underline{\mathcal M}\rho^j)$ ($j=1,\dots,\left\lceil {\log_{\rho} \overline{\mathcal M} / \underline{\mathcal M}}\right\rceil$, where $\lceil \cdot \rceil$ is the ceiling function). To speed up the \texttt{GetMax} operation, a variable $j_{max}$ keeps the index of the non-empty bucket containing the element with the largest margin. In the \texttt{Insert} operation, we simply compute the index $j$ of the corresponding bucket, insert the element in this bucket, and update $j_{max}$ if $j>j_{max}$. In the \texttt{GetMax} operation, we return \emph{any} element from the $j_{max}$-th bucket, and update $j_{max}$. Note that the margin of the returned element is not necessarily the maximum margin in the queue, but it is at least ${1}/{\rho }$ times this value. Since both operations (\texttt{Insert} and \texttt{GetMax}) can be carried out in $O(1)$, we have proved that \eqref{eq:LowerBoundInductive} can also be computed in $O(1)$.

Moreover, since the bounds in \eqref{eq:LowerBoundLocal} and \eqref{eq:UpperBoundLocal} are tighter if the regions involved are close to uniform (because in this case, given the summary, there is no uncertainty regarding the value of any point in the region), this choice of $\omega_k$ automatically drives the algorithm to focus on the edges of the image \emph{and} the prior, avoiding the need to subdivide and work on large uniform regions of the image \emph{or} the prior.

This concludes the derivation of the bounds to be used to solve our problem of interest. In the next section we show results obtained using these bounds integrated with the FoAM described in Section \ref{sec:FoAM}. 

\section{Experimental results}
\label{sec:ExperimentalResults}
In this section we apply the framework described in previous sections to the problem of simultaneously estimating the class, pose, and a denoised version (a segmentation) of a shape in an image. We start by analyzing the characteristics of the proposed algorithm on synthetic experiments (Section \ref{sec:SyntheticExperiments}), and then present experiments on real data (Section \ref{sec:RealExperiments}). These experiments were designed to test and illustrate the proposed theory only. Achieving state-of-the-art results \emph{for each} of the specific sub-problems would require further extensions of this theory.

\subsection{Synthetic experiments}
\label{sec:SyntheticExperiments}
\begin{figure}
\includegraphics[width=\columnwidth, bb=0pt 0pt 320pt 148pt]{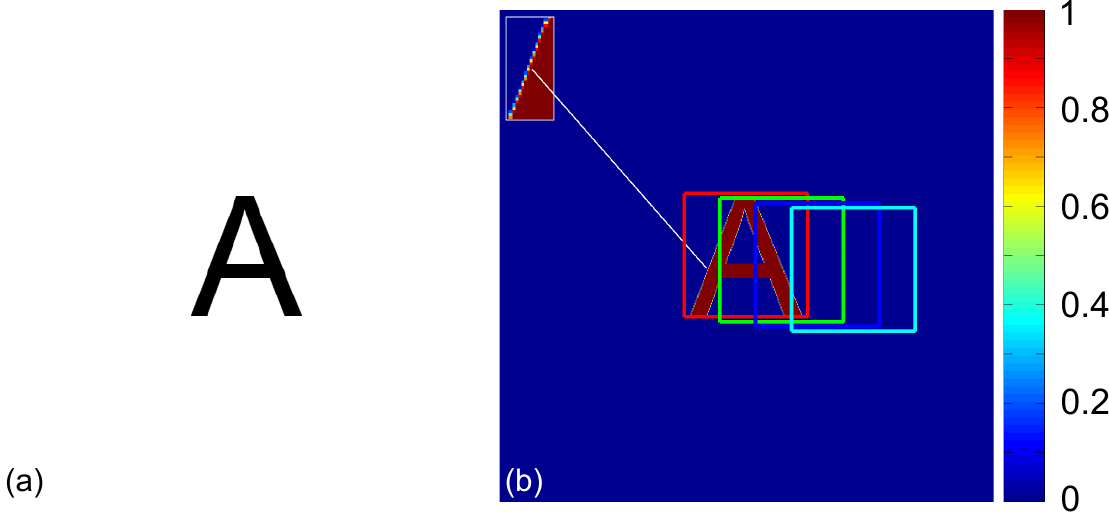}
\vspace{-10pt}
\caption{(a) The input image considered in Experiment 1 and (b) its corresponding BF with success rates $p_{B_f}$. (Inset) Zoom-in on the edge of the `A'. The rectangles in (b) indicate the supports corresponding to each of the 4 hypotheses defined in this experiment. The same colors are used for the same hypotheses in the following two figures. Throughout this article, use the colorbar on the right of this figure to interpret the colors of the BFs presented.}
\vspace{-15pt}
\label{fig:SetupFirstExperiment}
\end{figure}
In this section we present a series of synthetic experiments to expose the characteristics of the proposed approach. 

\paragraph{Experiment 1.}
We start with a simple experiment where both the input image (Fig. \ref{fig:SetupFirstExperiment}a) and the shape prior are constructed from a single shape (the letter `A'). Since we consider a single shape prior, we do not need to estimate the class of the shape in this case, only its pose. In this situation the success rates $p_{B_f}$ of the BF corresponding to this image (Fig. \ref{fig:SetupFirstExperiment}b), and the success rates $p_{B_K}$ of the BF corresponding to the shape prior, are related by a translation $\vec{t}$ (i.e., $p_{B_f}(\vec{x}) = p_{B_K}(\vec{x} - \vec{t})$). This translation is the \emph{pose} that we want to estimate. In order to estimate it, we define four hypotheses and use the proposed approach to select the hypothesis $H$ that maximizes the evidence $L(H)$. Each hypothesis is obtained for a different translation (Fig. \ref{fig:SetupFirstExperiment}b), but for the same shape prior.

\begin{figure}[b]
\begin{center}
\includegraphics[width=0.9\columnwidth, bb=0pt 0pt 1494pt 909pt]{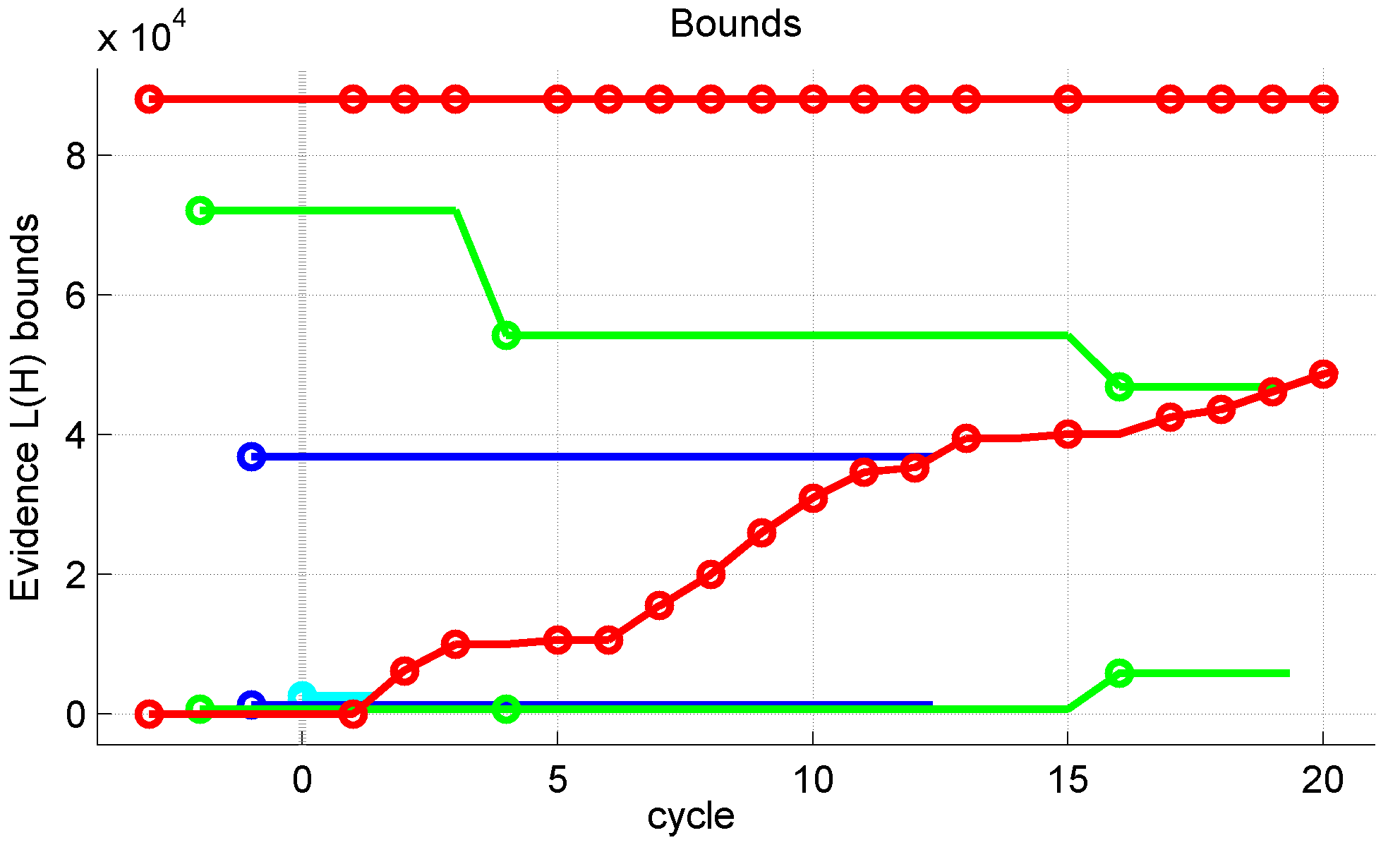}
\end{center}
\vspace{-15pt}
\caption{Progressive refinement of the evidence bounds obtained using a H\&B algorithm, for the four hypotheses defined in Fig. \ref{fig:SetupFirstExperiment}. The bounds of each hypothesis are represented by the two lines of the same color (the lower bound for the hypothesis in cyan, however, is occluded by the other lower bounds). During the initialization stage ($cycle \le 0$), the bounds of all hypotheses are initialized. Then, in the refinement stage ($cycle \ge 1$), one hypothesis is selected in each cycle and its bounds are refined (this hypothesis is indicated by the marker `o'). Hypotheses 4 (cyan), 3 (blue), and 2 (green) are discarded after the 2nd, 13th, and 19th refinement cycles, respectively, proving that Hypothesis 1 (red) is optimal. Note that it was not necessary to compute the evidence exactly to select the best hypothesis: The bounds are sufficient and much cheaper to compute.}
\label{fig:EvolutionOfBounds}
\end{figure}

\begin{figure}
\begin{center}
\includegraphics[width=\columnwidth, bb=0pt 0pt 1950pt 575pt]{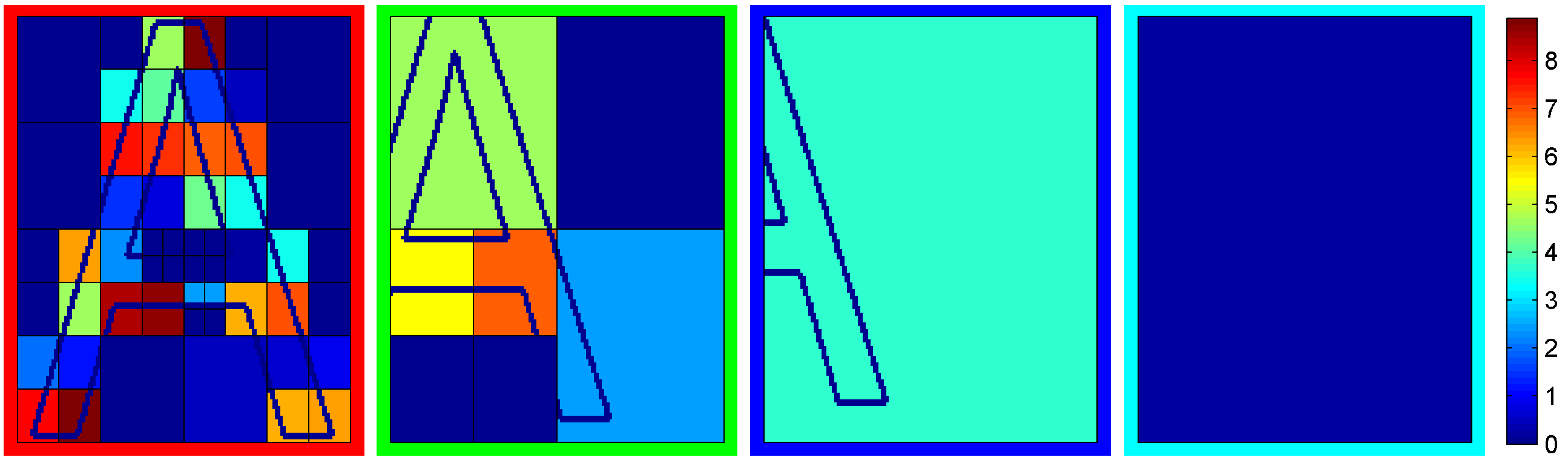}
\end{center}
\vspace{-10pt}
\caption{Final partitions obtained for the four hypotheses in Experiment 1. The color of each element of the partition (i.e., rectangle) indicates the \emph{margin} of the element (use the colorbar on the right to interpret these colors). Note that higher margins are obtained for the elements around the edges of the image or the prior, and thus these areas are given priority during the refinements. The edges of the letter `A' were included for reference only.}
\label{fig:FinalMargins}
\vspace{-15pt}
\end{figure}

As described in Section \ref{sec:FoAM}, at the beginning of the algorithm the bounds of all hypotheses are initialized, and then during each iteration of the algorithm one hypothesis is selected, and its bounds are refined (Fig. \ref{fig:EvolutionOfBounds}). It can be seen in Fig. \ref{fig:EvolutionOfBounds} that the FoAM allocates more computational cycles to refine the bounds of the best hypothesis (in red), and less cycles to the other hypotheses. In particular, two hypotheses (in cyan and blue) are discarded after spending just one cycle (the initialization cycle) on each of them. Consequently, it can be seen in Fig. \ref{fig:FinalMargins} that the final partition for the red hypothesis is finer than those for the othe hypotheses. It can also be seen in the figure that the partitions are preferentially refined around the edges of the image or the prior (remember from \eqref{eq:L_H} that image and prior play a symmetric role), because the partition elements around these areas have greater margins. In other words, the FoAM algorithm is ``paying attention" to the edges, a sensible thing to do. Furthermore, this behavior was not intentionally ``coded" into the algorithm, but rather it emerged as the BM greedily minimizes the margin.

To select the best hypothesis in this experiment, the functions to compute the lower and upper bounds (i.e., those that implement \eqref{eq:LowerBoundLocal}-\eqref{eq:DiscreteShapeLower} and \eqref{eq:UpperBoundLocal}-\eqref{eq:SemiDiscreteShapeUpper}) were ca\-lled a total of 88 times each. In other words, 22 pairs of bounds were computed, on average, for each hypothesis. In contrast, if $L(H)$ were to be computed exactly, 16,384 pixels would have to be inspected for each hypothesis (because all the priors used in this section were of size $128 \times 128$). While inspecting one pixel is significantly cheaper than computing one pair of bounds, for images of ``sufficient" resolution the proposed approach is more efficient than na\"{i}vely inspecting every pixel. Since the relative cost of evaluating one pair of bounds (relative to the cost of inspecting one pixel) depends on the implementation, and since at this point an efficient implementation of the algorithm is not available, we use the \emph{average number of bound pairs evaluated per hypothesis} (referred as $\tau$) as a measure of performance (i.e., $\tau \triangleq \text{Pairs of bounds computed} / \text{Number of Hypotheses}$).

Moreover, for images of sufficient resolution, not only is the amount of computation required by the proposed algorithm less than that required by the na\"{i}ve approach, it is also \emph{independent of the resolution} of these images. For example, if in the previous experiment the resolution of the input image and the prior were doubled, the number of pixels to be processed by the na\"{i}ve approach would increase four times, while the number of bound evaluations would remain the same. In other words, \emph{the amount of computation needed to solve a particular problem using the proposed approach only depends on the problem, not on the resolution of the input image}. 

\paragraph{Experiment 2.}
The next experiment is identical to the previous one, except that one hypothesis is defined for every possible integer translation that yields a hypothesis whose support is contained within the input image. This results in a total of 148,225 hypotheses. In this case, the set $\mathbb{A}$ of active hypotheses when termination conditions were reached contained 3 hypotheses. We refer to this set as the \emph{set of solutions}, and to each hypothesis in this set as a \emph{solution}. Note that having reached termination conditions with a set of solutions having more than one hypothesis (i.e., solution) implies that all the hypotheses in this set have been completely refined (i.e., either $p_{B_f}$ or $p_{B_H}$ are uniform in all their partition elements). 

To characterize the set of solutions $\mathbb{A}$, we define the \emph{translation bias}, $\mu_{\vec{t}}$, and the \emph{translation standard deviation}, $\sigma_{\vec{t}}$, as
\begin{align}
\mu_{\vec{t}} & \triangleq \left \| \frac{1}{|\mathbb{A}|} \sum_{i=1}^{|\mathbb{A}|}{\vec{t}_i - \vec{t}_T} \right \|, \ \text{and} \\
\sigma_{\vec{t}} & \triangleq \sqrt{ \frac{1}{|\mathbb{A}|} \sum_{i=1}^{|\mathbb{A}|}{\left \| \vec{t}_i - \vec{t}_T \right \|^2}},
\end{align} 
respectively, where $\vec{t}_i$ is the translation corresponding to the $i$-th hypothesis in the set $\mathbb{A}$ and $\vec{t}_T$ is the true translation. In this particular experiment we obtained $\mu_{\vec{t}}=0$ and $\sigma_{\vec{t}}=0.82$, and the set $\mathbb{A}$ consisted on the true hypothesis and the two hypotheses that are one pixel translated to the left and right. These 3 hypotheses are indistinguishable under the conditions of the experiment. There are two facts contributing to the uncertainty that makes these hypotheses indistinguishable: 1) the fact that the edges of the shape in the image and the prior are not sharp (i.e., not having probabilities of 0/1, see inset in Fig. \ref{fig:SetupFirstExperiment}b); and 2) the fact that $m < \infty$ in the $m$-summaries and hence some ``information" of the LCDF is ``lost" in the summary, making the bounds looser.

\begin{figure}
\vspace{-5pt}
\includegraphics[width=\columnwidth, bb=0pt 0pt 3174pt 1274pt]{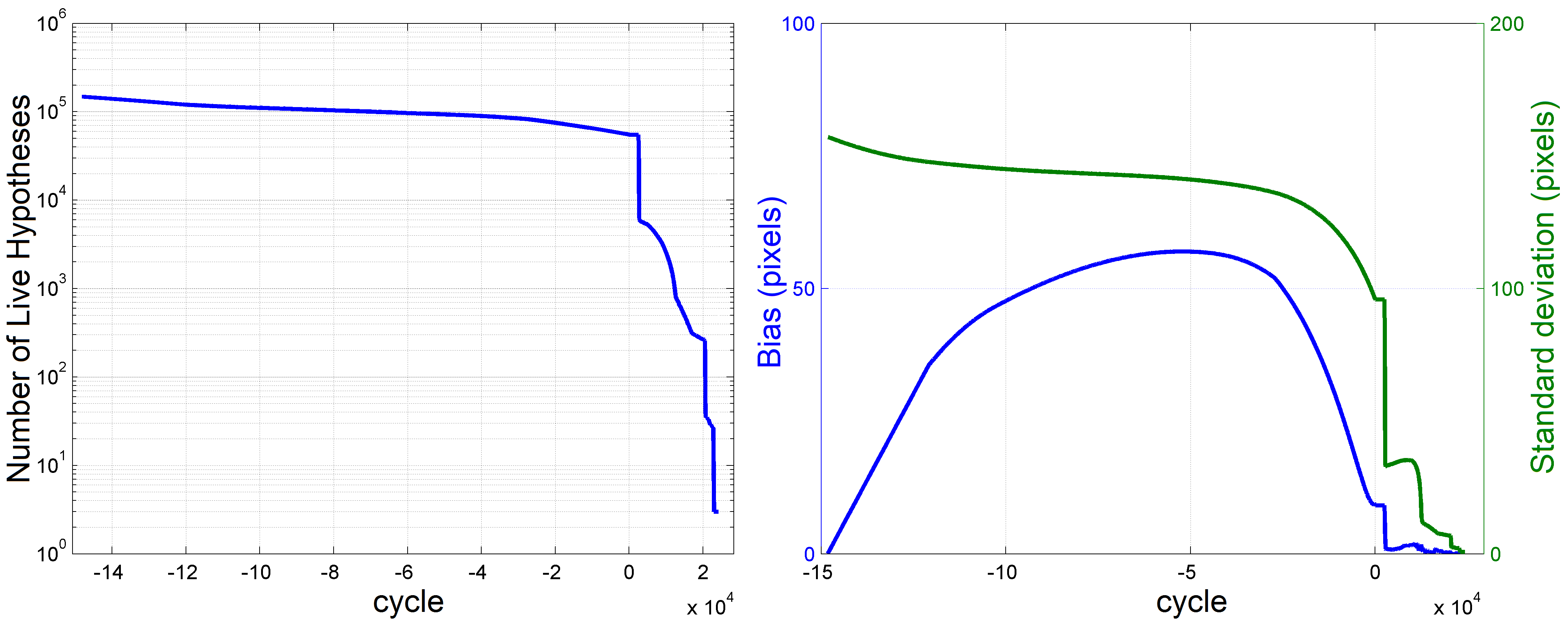}
\vspace{-10pt}
\caption{(Left) Evolution of the number of live hypotheses ($|\mathbb{A}|$) as a function of time. (Right) Evolution of the translation bias $\mu_{\vec{t}}$ (blue) and standard deviation $\sigma_{\vec{t}}$ (green) of the set of active hypotheses $\mathbb{A}$.}
\label{fig:Exp2Evolutions}
\vspace{-5pt}
\end{figure}

Figure \ref{fig:Exp2Evolutions} shows the evolution of the set $\mathbb{A}$ of active hypotheses as the bounds get refined. Observe in this figure that during the refinement stage of the algorithm ($cycles > 0$ in the figure), the number of active hypotheses ($|\mathbb{A}|$), the bias $\mu_{\vec{t}}$, and the standard deviation $\sigma_{\vec{t}}$ sharply decrease. It is  interesting to note that during the first half of the initialization stage, because hypotheses are not discarded symmetrically around the true hypothesis, the bias increases. 

\begin{figure} \sidecaption
\includegraphics[width=0.57\columnwidth, bb=0pt 0pt 2316pt 2234pt]{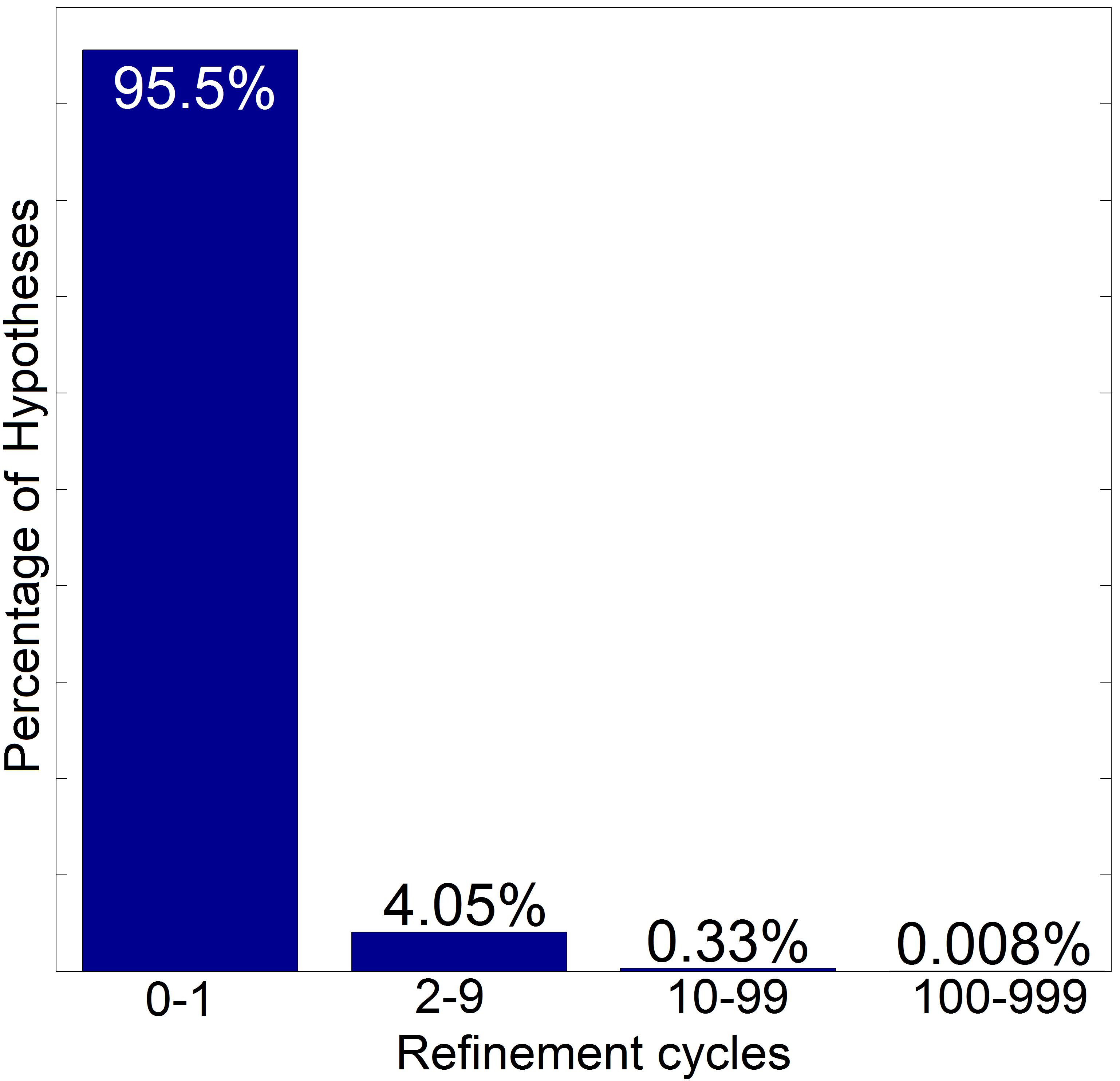}
\vspace{-10pt}
\caption{Percentage of the hypotheses that were refined a certain number of times. Most hypotheses (95.5\%) were refined only once, or not refined at all.}
\label{fig:Exp2Histogram}
\end{figure}

Figure \ref{fig:Exp2Histogram} shows the percentage of hypotheses that were refined 0 or 1 times, between 2 and 9 times, between 10 and 99 times, and between 100 and 999 times. This figure indicates that, as desired, \emph{very little computation is spent on most hypotheses, and most computation is spent on very few hypotheses}. Concretely, the figure shows that for 95.5\% of the hypotheses, either the initialization cycle is enough to discard the hypothesis (i.e., only 1 pair of bounds needs to be computed), or an additional refinement cycle is necessary (and hence $1+4=5$ pairs of bounds are computed).  On the other hand, only 0.008\% of the hypotheses require between 100 and 999 refinement cycles. On average, only 1.78 pairs of bounds are computed for each hypothesis ($\tau = 1.78$), instead of inspecting 16,384 pixels for each hypothesis as in the na\"{i}ve approach. For convenience, these results are summarized in Table \ref{tab:Exp2Results}.

\begin{table}[h]
\vspace{-10pt}
\caption{Estimation of the position of a known shape `A' in a noiseless input image.}
\label{tab:Exp2Results}
\begin{tabular*}{\columnwidth}{@{\extracolsep{\fill}} | c | c | c | c |} \hline 
$|\mathbb{A}|$ & $\mu_{\vec{t}} (pixels)$ & $\sigma_{\vec{t}}$(pixels) & $\tau$ \\ \hline 
 3 & 0 & 0.82 & 1.78 \\ \hline
 \end{tabular*}
\vspace{-10pt}
\end{table}

\paragraph{Experiment 3.}
In the next experiment the hypothesis space is enlarged by considering not only integer translations, but also scalings. These scalings changed the horizontal and vertical size of the prior (independently in each direction) by multiples of 2\%. In other terms, the transformations used were of the form 
\begin{equation}
T(\vec{x}) = Diag(\vec{s}) \vec{x} + \vec{t},
\label{eq:Transformation}
\end{equation}
where $\vec{t} \triangleq [t_x,t_y]$ is an integer translation (i.e., $t_x, t_y \in \mathbb{Z}$), $Diag(\vec{s})$ is a diagonal matrix containing the vector of scaling factors $\vec{s} \triangleq [s_x s_y]$ in its diagonal, and $s_x, s_y \in \{0.96, 0.98, 1, 1.02,$ $1.04 \}$.

To characterize the set of solutions $\mathbb{A}$ of this experiment, in addition to the quantities $\mu_{\vec{t}}$ and $\sigma_{\vec{t}}$ that were defined before, we also define the \emph{scaling bias}, $\mu_{\vec{s}}$, and the \emph{scaling standard deviation}, $\sigma_{\vec{s}}$, as
\begin{align}
\mu_{\vec{s}} & \triangleq \left \| \frac{1}{|\mathbb{A}|} \sum_{i=1}^{|\mathbb{A}|}{\vec{s}_i - \mathbf{1}} \right \|, \ \text{and} \\
\sigma_{\vec{s}} & \triangleq \sqrt{ \frac{1}{|\mathbb{A}|} \sum_{i=1}^{|\mathbb{A}|}{\left \| \vec{s}_i - \mathbf{1} \right \|^2}},
\end{align} 
respectively, where $\vec{s}_i$ are the scaling factors corresponding to the $i$-th hypothesis in the set $\mathbb{A}$ (the true scaling factors, for simplicity, are assumed to be $s_x=s_y=1$).

In this experiment the set of solutions consisted of the same three hypotheses found in Experiment 2, plus two more hypotheses that had a $\pm2\%$ scaling error. The performance of the framework in this case, on the other hand, improved with respect to the previous experiment (note the $\tau$ in Table \ref{tab:ExpScale}). These results (summarized in Table \ref{tab:ExpScale}) suggest that in addition to the translation part of the pose, the scale can also  be estimated very accurately and efficiently.

\begin{table}[h]
\vspace{-10pt}
\caption{Errors in the estimation of the position and scale of a \emph{known} shape `A' in a noiseless input image.}
\label{tab:ExpScale}
\begin{tabular*}{\columnwidth}{@{\extracolsep{\fill}} | c | c | c | c | c | c |} \hline 
$|\mathbb{A}|$ & $\mu_{\vec{t}}$ (pixels) &   $\sigma_{\vec{t}}$ (pixels)   &  $\mu_S$ (\%) & $\sigma_S$ (\%)  & $\tau$ \\ \hline 
5 & 0 & 0.633 & 0 & 1.26 & 1.17 \\ \hline
 \end{tabular*}
\vspace{-10pt}
\end{table}

\begin{figure}
\vspace{-5pt}
\begin{center}
\includegraphics[width=0.8\columnwidth, bb=0pt 0pt 465pt 775pt]{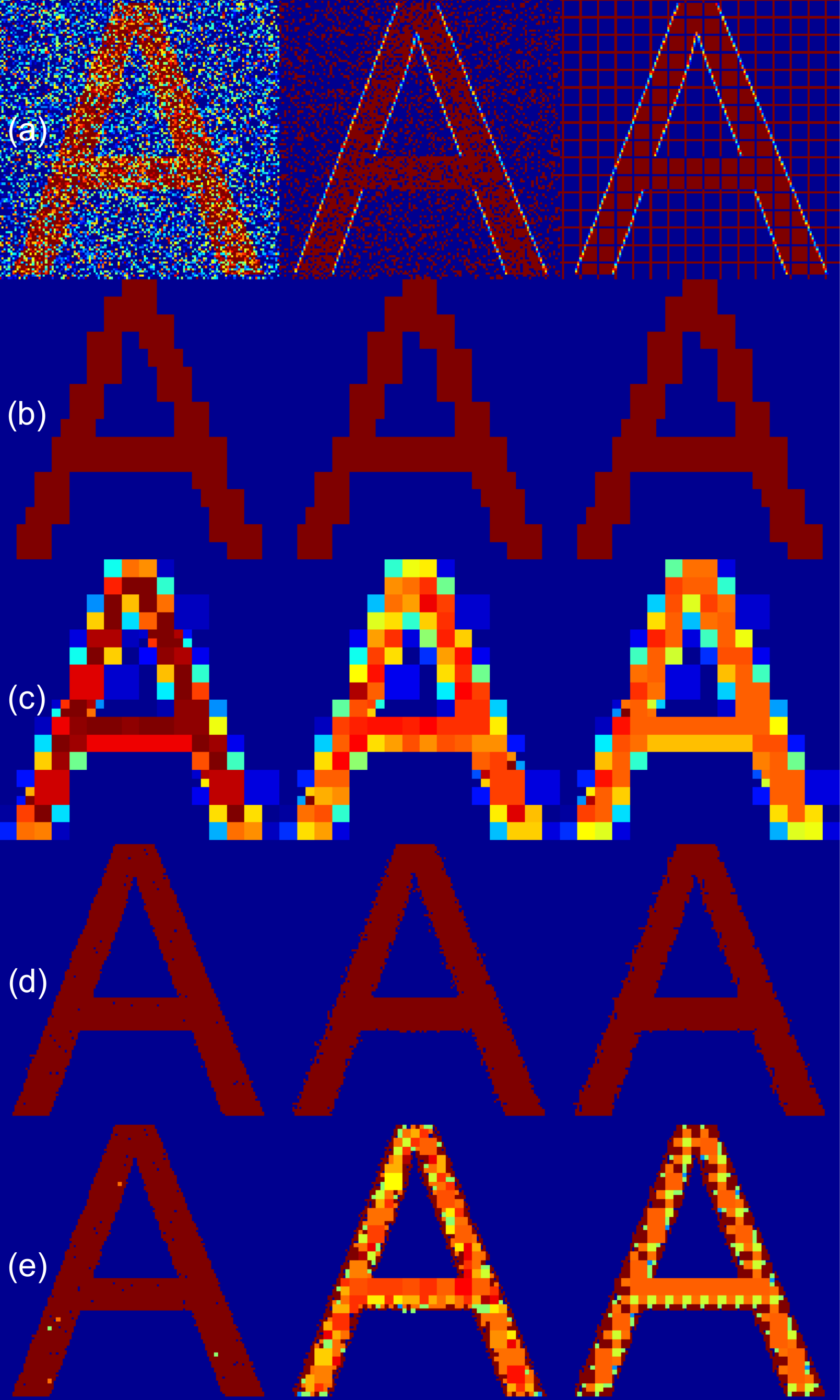}
\end{center}
\vspace{-10pt}
\caption{(a) BFs corresponding to the input image after being degraded by three different kinds of noise (from left to right): Gaussian noise $\mathcal{N}(0, 0.45^2)$; Salt and pepper noise $\mathcal{SP}(0.25)$; and structured noise $\mathcal{S}(8)$. (b-c) Discrete shape $\hat{S}$ (b), and semidiscrete shape $\tilde{S}$ (c), estimated for the best solution after 50 refinement cycles. Each shape was obtained for the corresponding input image depicted in the same column in (a). (d-e) Discrete (d) and semidiscrete (e) shapes obtained when termination conditions were achieved, for each of the input images in (a). The colors of the semidiscrete shapes at each partition element $\omega$ represent the fraction of the element (i.e., $\tilde{S}(\omega) / |\omega|$) that is covered by the shape (use the colorbar in Fig. \ref{fig:SetupFirstExperiment} to interpret these colors).}
\label{fig:Exp3Noise}
\vspace{-15pt}
\end{figure}

\paragraph{Experiment 4.}
The performance of the framework is obviously affected when the input image is corrupted by noise. To understand how it is affected, we run the proposed approach with the same hypothesis space defined in Experiment 2, but on images degraded by different kinds of noise (for simplicity we add the noise directly to the BF corresponding to the input image, rather than to the input image itself). Three kinds of noise have been considered (Fig. \ref{fig:Exp3Noise}a): 1) additive, zero mean, white Gaussian noise with standard deviation $\sigma$, denoted by $\mathcal{N}(0,\sigma^2)$; salt and pepper noise, $\mathcal{SP}(P)$, produced by transforming, with probability $P$, the success rate $p(\vec{x})$ of a pixel $\vec{x}$ into $1-p(\vec{x})$; and structured noise, $\mathcal{S}(\ell)$, produced by transforming the success rate $p(\vec{x})$ into $1-p(\vec{x})$ for each pixel $\vec{x}$ in rows and colums that are multiples of $\ell$. When adding Gaussian noise to a BF some values end up outside the interval $[0,1]$. In such cases we trim these values to the corresponding extreme of the interval. The results of these experiments are summarized in Table \ref{tab:EffectOfNoise}.

\begin{table}[h]
\caption{Effect of corrupting the BF corresponding to the input image with noise.}
\label{tab:EffectOfNoise}
\begin{tabular*}{\columnwidth} {@{\extracolsep{\fill}} | c | c | c | c | c |} \hline 
Noise  &  $|\mathbb{A}|$  &  $\mu_{\vec{t}} (pixels)$  &  $\sigma_{\vec{t}}$(pixels)  &  $\tau$  \\ \hline 

$\mathcal{N}(0, 0.15^2)$ & 12 & 0.16 & 1.41 & 4.59 \\ \hline
$\mathcal{N}(0, 0.30^2)$ & 11 & 0 & 1.34 & 2.66 \\ \hline
$\mathcal{N}(0, 0.45^2)$ & 11 & 0 & 1.34 & 3.12 \\ \noalign{\hrule height 1pt}
$\mathcal{SP}(0.083)$ & 3 & 0 & 0.81 & 1.6 \\ \hline
$\mathcal{S}(24)$ & 6 & 0.5 & 1.08 & 2.76 \\ \noalign{\hrule height 1pt}
$\mathcal{SP}(0.125)$ & 3 & 0 & 0.81 & 6.28 \\ \hline
$\mathcal{S}(16)$ & 3 & 0 & 0.81 & 5.49 \\ \noalign{\hrule height 1pt}
$\mathcal{SP}(0.25)$ & 3 & 0.47 & 1.15 & 42.77 \\ \hline
$\mathcal{S}(8)$ & 5 & 0.4 & 1.09 & 39.17 \\ \hline

 \end{tabular*}
\end{table}

By comparing tables \ref{tab:Exp2Results} and \ref{tab:EffectOfNoise} we observe that the approach is relatively immune to the levels and kinds of noise applied, since $\mu_{\vec{t}}$ and $\sigma_{\vec{t}}$ did not increase much. However, the effects of the different kinds of noise are different. Adding moderate amounts of Gaussian noise, by making it uncertain whether each pixel in the image belongs to the foreground or background, increased by more than 3 times the size of the set of solutions, while doubling the amount of computation required (compare $\tau$ in tables \ref{tab:Exp2Results} and \ref{tab:EffectOfNoise}). In contrast, moderate amounts of salt and pepper noise almost did not affect the set of solutions found, but they dramatically increased the necessary computation (because this kind of noise created many more ``edges" in the image). To understand the effect of the position of the errors introduced by the noise, we adjusted the level of the structured noise $\ell$ so that the same (approximate) number of pixels were affected by it and by the salt and pepper noise (i.e., approximately the same number of pixels were affected by $\mathcal{SP}(0.083)$ and $\mathcal{S}(24)$, by $\mathcal{SP}(0.125)$ and $\mathcal{S}(16)$, and by $\mathcal{SP}(0.25)$ and $\mathcal{S}(8)$). Because the structured noise concentrates the ``errors" in some parts of the image, this kind of noise increased slightly the size of the set of solutions, but it did not have a consistent effect on the errors or the amount of computation required.

Fig. \ref{fig:Exp3Noise}b-e show the shapes that were estimated from the noisy BFs in Fig. \ref{fig:Exp3Noise}a. To understand how these shapes were estimated, recall that a discrete shape (given by \eqref{eq:DiscreteShapeLower}) is implicitly obtained while computing the lower bound of a hypothesis, and a semidiscrete shape (given by \eqref{eq:SemiDiscreteShapeUpper}) is implicitly obtained while computing the upper bound of a hypothesis. Thus, Fig. \ref{fig:Exp3Noise}d-e show the discrete and semidiscrete shapes obtained for the solution with the greatest upper bound. Notice how we were able to remove almost all the noise. It is also interesting to inspect the shapes estimated at an intermediate stage of the process, after only 50 refinement cycles of the bounds had been performed (Fig. \ref{fig:Exp3Noise}b-c).

\paragraph{Experiment 5.}
Another factor that affects the framework's performance is the ``shape" of the prior. As mentioned before (in Section \ref{sec:IncrementalBounds}), the bounds computed for a partition element are tightest when either one of the BFs corresponding to the prior or the image are uniform in the element. This is because when one of the BFs is uniform, given the \emph{m}-summary, there is no uncertainty about the location of the values of $\delta(\vec{x})$ in the element. Due to this fact, shapes that are uniform on elements of a coarser partition (referred below as \emph{``coarser" shapes}) are processed faster than shapes that are uniform only on the elements of a finer partition (referred below as \emph{``finer" shapes}). This statement is quantified in Table \ref{tab:EffectOfShape} which summarizes the results of five experiments.
\begin{table}[h]
\vspace{-10pt}
\caption{Effect of the shape on the framework's performance.}
\label{tab:EffectOfShape}
\begin{tabular*}{\columnwidth}{@{\extracolsep{\fill}} | c | c | c | c | c | } \hline 
Shape  &   $|\mathbb{A}|$   &   $\mu_{\vec{t}} (pixels)$   &   $\sigma_{\vec{t}}$(pixels)   &   $\tau$  \\ \hline 

$\blacksquare$ & 1 & 0 & 0 & 1.00 \\ \hline
$\CIRCLE$ & 1 & 0 & 0 & 1.04 \\ \hline
$\blacktriangle$ & 1 & 0 & 0 & 1.10 \\ \hline
$\clubsuit$ & 1 & 0 & 0 & 1.50 \\ \hline
$\bigstar$ & 1 & 0 & 0 & 1.77 \\ \hline

\end{tabular*}
\vspace{-20pt}
\end{table}

Each of these experiments is identical to Experiment 2, except that a different shape was used instead of the shape `A'. These shapes are depicted on the first column of Table \ref{tab:EffectOfShape}. Note that changing the shape in these experiments did not affect the set of solutions found (in all cases this set contains only the true solution and thus $\mu_{\vec{t}}=\sigma_{\vec{t}}=0$), but it did affect the amount of computation required in each case (see the column labeled `$\tau$'). In particular, note that since the shapes are roughly sorted from ``coarser" to ``finer" (as defined above),  the total computation required correspondingly increases. Interestingly, to estimate the pose of the first shape (a square of size $128 \times 128$), \emph{only one pair of bounds were computed} (in constant time) per hypothesis (instead of processing the 16,384 pixels of the shape prior).

\subsection{Experiments on real data}
\label{sec:RealExperiments}
In this section we apply the proposed framework to the problem of estimating the class, pose, and segmentation of a shape (a letter in this case), for each image in a set of \emph{testing} images. In order to do this, as before, we define multiple hypotheses and use the proposed framework to find the group of hypotheses that best explains each test image.

More specifically, one hypothesis is defined for each transformation (described by \eqref{eq:Transformation}) and for each class. To save memory, and since it was shown before that hypotheses that are far from the ground truth are easily discarded, we only considered integer translations (around the ground truth) of up to 5 pixels in each direction (i.e., $t_x,t_y \in \{-5,-4, \dots, 5\}$), and scalings (around the ground truth) by multiples of 4\% (i.e., $s_x,s_y \in \{0.92, 0.96, \dots, 1.08\}$). In contrast with the experiments presented in the previous section where only one class was considered (and hence there was no need to estimate it), in this section multiple classes are considered and estimating the class of the shape is part of the problem. We consider 6 different priors for each letter of the English alphabet, giving a total of $26 \times 6=156$ different priors (and classes). This results in an initial set of hypotheses containing 471,900 hypotheses (121 translations $\times$ 25 scalings $\times$ 156 priors).

\begin{figure}
\vspace{-5pt}
\includegraphics[width=\columnwidth, bb=0pt 0pt 2228pt 807pt]{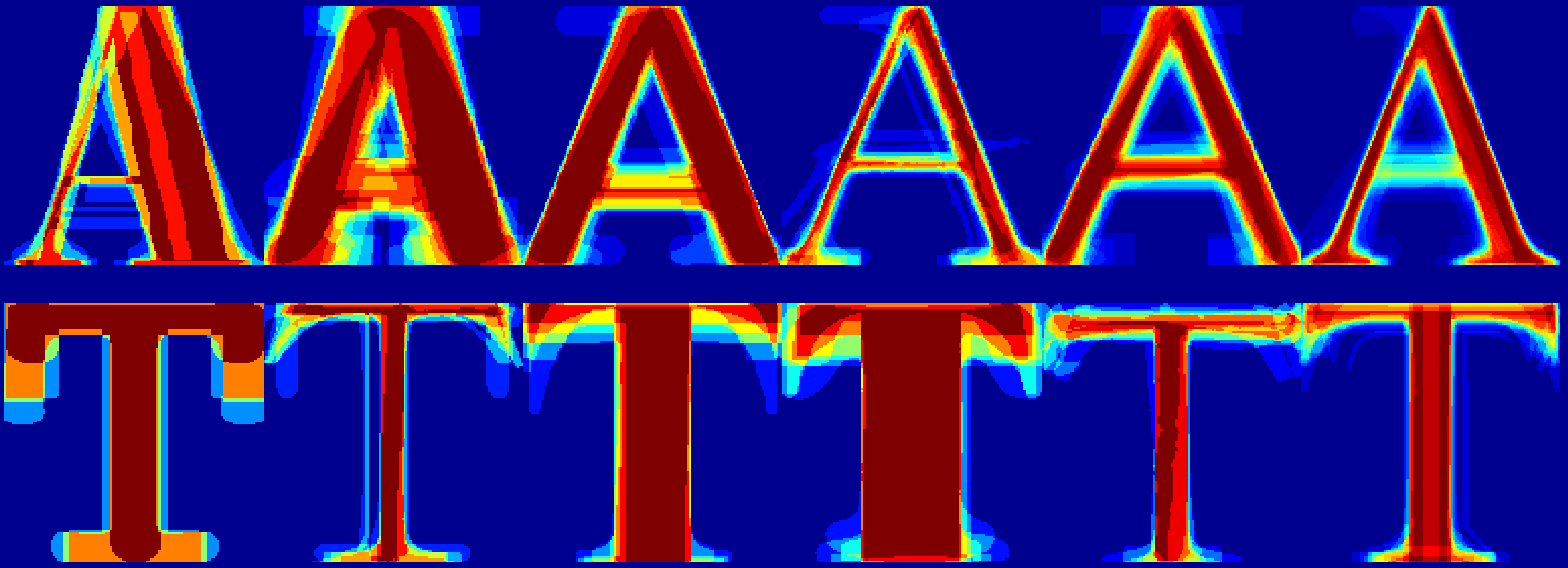}
\vspace{-10pt}
\caption{The six shape priors (BFs) obtained for each of the classes `A' and `T'.}
\label{fig:RealExpClassPriors}
\vspace{-15pt}
\end{figure}

To construct the priors for each letter, we compiled a set of \emph{training} shapes from the standard set of fonts in the Windows operating system. This set was pruned by discarding italicized fonts and other fonts that were considered outliers (e.g., wingdings). The remaining shapes were then divided into six subsets using \emph{k}-medoids clustering \cite{Has09}, where the distance $d(S_1,$ $S_2)$ between two shapes $S_1$ and $S_2$ was defined as the number of different pixels between the two shapes (i.e., $d(S_1,S_2) \triangleq |S_1 \cup S_2 \setminus S_1 \cap S_2|$) after ``proper alignment" (this ``alignment" is described below). It can be shown that in the case of 0/1 priors (``shapes"), this distance is equivalent to the evidence defined in  \eqref{eq:L_H}. Then all the letters in each cluster were aligned with respect to the medoid of the cluster by finding the optimal translation and scaling that minimizes the distance $d$ mentioned above, and \eqref{eq:BFFromTrainingSet} was used to compute the prior corresponding to the cluster. The six priors corresponding to the letters `A' and `T' are shown in Fig. \ref{fig:RealExpClassPriors}. The number of priors per class (six) was chosen to maximize the classification performance $P_1$ (defined below). We also tested having a variable number of clusters per class, but doing so did not improve the classification performance. The resulting priors used in the following experiments, as well as the set of training shapes used to construct them, can be downloaded from \cite{SupMat1}.

\begin{figure}
\vspace{-5pt}
\includegraphics[width=\columnwidth, bb=0pt 0pt 2625pt 970pt]{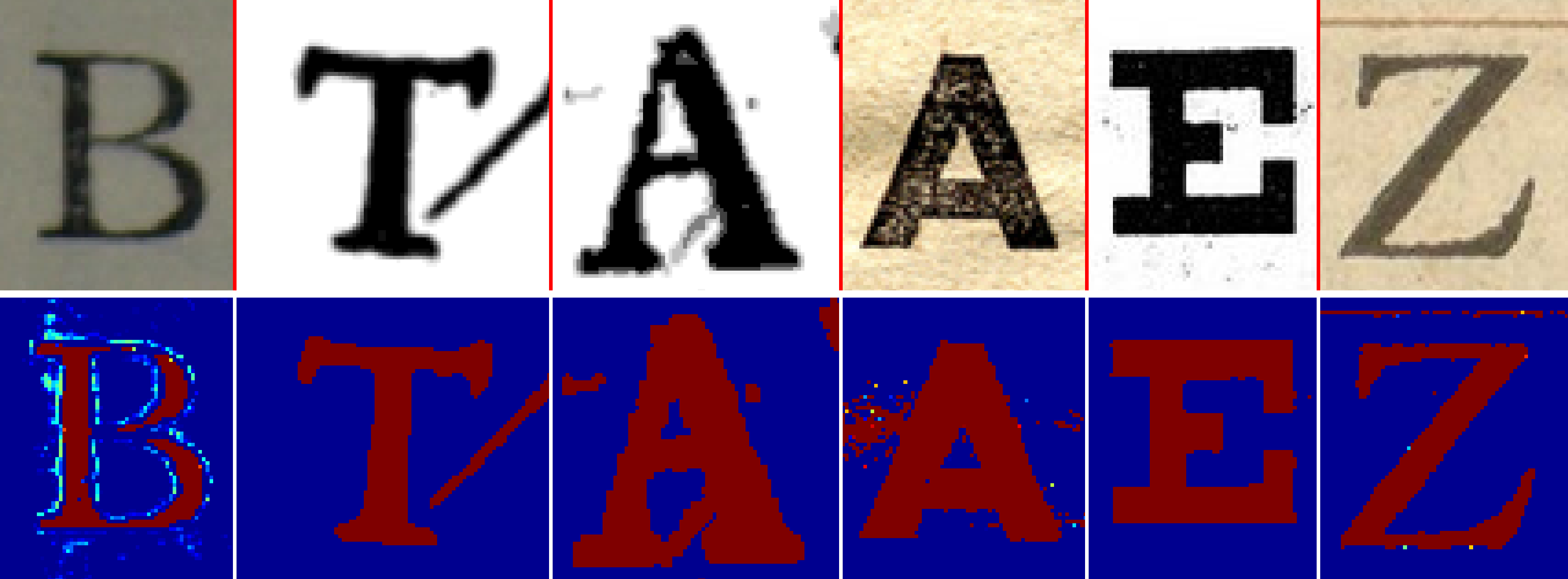}
\vspace{-10pt}
\caption{Sample letters in the testing set (top) and their corresponding BFs (bottom), obtained by background subtraction using \eqref{eq:BFBackgrounSubtraction}. Note the different types of noise present and the different typographies.}
\label{fig:RealExpDataset}
\end{figure}

The testing images, on the other hand, were obtained by extracting rectangular regions (containing each a single letter) from old texts exhibiting decoloration and ink bleedthough, among other types of noise (see examples in Fig. \ref{fig:RealExpDataset} and in the images of these old texts in \cite{SupMat1}). The corresponding BFs were obtained from these images by background subtraction using \eqref{eq:BFBackgrounSubtraction}. The probability density functions (pdf's) $p(f|q=0)$ and $p(f|q=1)$ in that equation were learned from sets of pixels in the background and foreground that were manually selected for each image. In this case we learned a single pdf for the background, and a single pdf for the foreground (not a different pdf for each pixel). While more sophisticated methods could have been used to obtain these BFs, this is not the goal of this article, and in fact, noisy BFs are desirable to test the robustness of the approach. This testing set, containing 218 letters, is used to measure the classification performance of the proposed framework. It is important to note that the testing and training sets come from \emph{different} sources (i.e., there is no overlap between the testing and training sets). 

\begin{figure}
\vspace{-5pt}
\begin{center}
\includegraphics[width=\columnwidth, bb=0pt 0pt 2928pt 997pt]{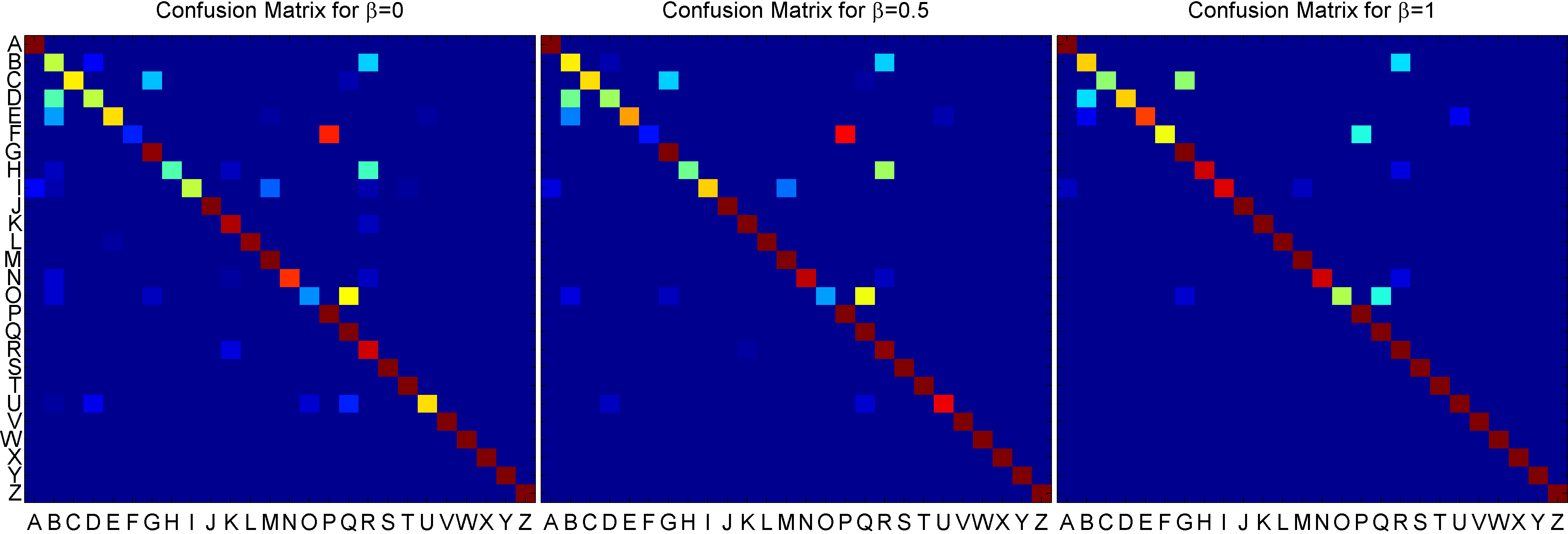}
\end{center}
\vspace{-10pt}
\caption{Confusion matrices for $\beta=0$ (left), $\beta=0.5$ (middle), and $\beta=1$ (right). Use the colorbar in Fig. \ref{fig:SetupFirstExperiment} to interpret the colors.}
\label{fig:ConfusionMatrices}
\vspace{-17pt}
\end{figure}

Since the proposed approach does not necessarily associate a single class to each testing image (because the set of solutions obtained for an image might contain solutions of different classes), we report the performance of the framework with a slight modification of the traditional indicators. An element $(i,j)$ in the \emph{confusion matrix} $C_0$ (in Fig. \ref{fig:ConfusionMatrices}) indicates the fraction of solutions of class $j$ (among \emph{all} the solutions) obtained for all testing images of class $i$. It is also of interest to know what the classification performance is when only the best solutions are considered. For this purpose we define the confusion matrix $C_\beta$ ($0 \le \beta \le 1$) as before, but considering only the solutions whose upper bound is greater or equal than $\gamma_\beta$, where $\gamma_\beta \triangleq \underline{L} + \beta (\overline{L} - \underline{L})$ and $\underline{L}$ and $\overline{L}$ are the maximum lower and upper bounds, respectively, of any solution. Note that $\gamma_0=\gamma=\underline{L}$ when $\gamma$ is defined as in Section \ref{sec:FoAM}, and hence all the solutions are considered, and that $\gamma_1=\overline{L}$ and hence only the solution with the largest upper bound is considered. The confusion matrices $C_{0.5}$ and $C_1$ are also shown in Fig. \ref{fig:ConfusionMatrices}.

Similarly, the total classification performance, $P_\beta$, is defined as the number of solutions of the correct class (accumulated over all the images in the testing set), divided by the total number of solutions. As above, the solutions considered are only those whose upper bound is greater or equal than $\gamma_\beta$. The total performance obtained in this case was $P_0=82.5\%$, $P_{0.5}=86.5\%$ and $P_1=90.4\%$. As expected, when only the best solutions are considered, the performance improves (i.e., $P_1 \ge P_{0.5} \ge P_0$).

\begin{figure}
\includegraphics[width=\columnwidth, bb=0pt 0pt 1747pt 584pt]{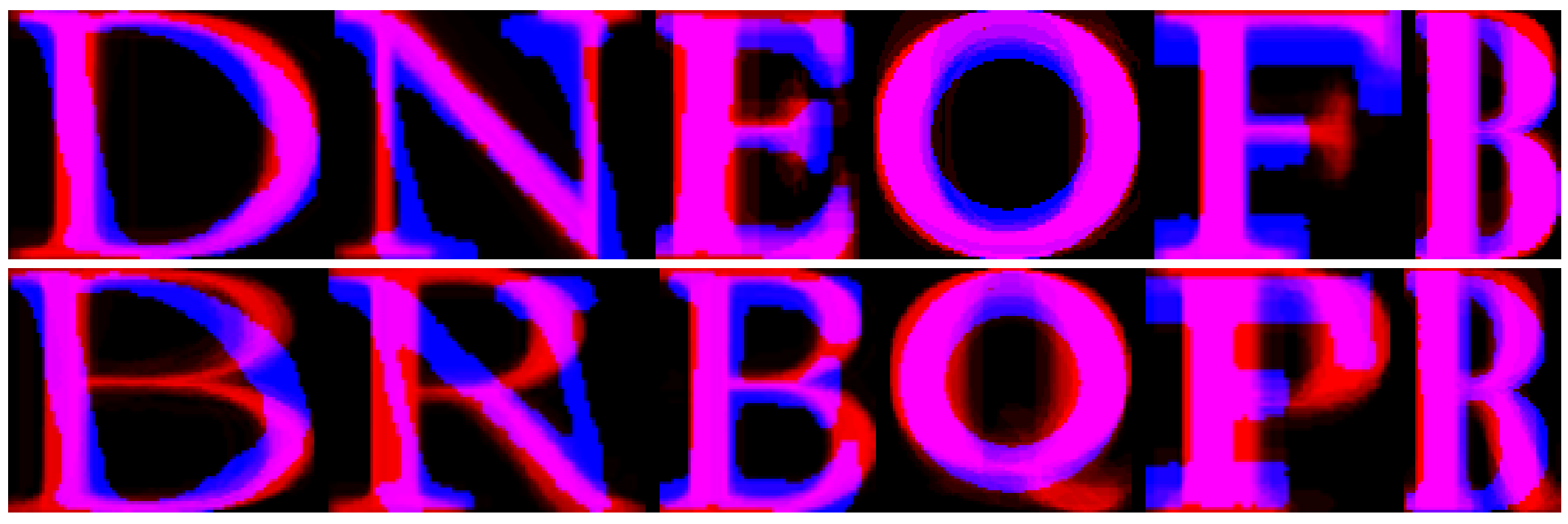}
\vspace{-10pt}
\caption{Representative examples of the solutions found by the proposed framework, given by their corresponding BFs. Solutions in the top row are of the correct class, while solutions in the bottom row are of a different class. These composite images are produced by overlaying the BF corresponding to the testing image (in the blue channel) and the BF corresponding to the solution (in the red channel).}
\label{fig:TypicalResults}
\vspace{-15pt}
\end{figure}

For completeness, Table \ref{tab:ExpTestingDataset} summarizes the \emph{average} pose estimation ``errors" obtained in this case. The quantities $\overline{|\mathbb{A}|}$, $\overline{\mu_{\vec{t}}}$, $\overline{\sigma_{\vec{t}}}$, $\overline{\mu_S}$, and $\overline{\sigma_S}$ in this table were respectively obtained as the average of the quantities $|\mathbb{A}|$, $\mu_{\vec{t}}$, $\sigma_{\vec{t}}$, $\mu_S$, and $\sigma_S$ (defined before) over all images in the testing set. In contrast with the synthetic experiments reported in Section \ref{sec:SyntheticExperiments}, for the experiments reported in this section the ground truth pose is not available. For this reason, and for lack of a universally trusted definition of the \emph{true pose} (which is ill defined when the shapes to be aligned are different and unknown), in these experiments the ground truth pose was defined by the center (for the location) and the size (for the scale) of a manually defined bounding box around each letter. Therefore, the errors in Table \ref{tab:ExpTestingDataset} must be analyzed with caution, since it is not clear that this arbitrary defintion of the ground truth should be preferred over the solutions returned by the framework. In fact, in all the inspected cases (e.g., see Fig. \ref{fig:TypicalResults}), we did not observe a clear ``misalignment."

\begin{table}[h]
\vspace{-10pt}
\caption{Mean ``errors" in the estimation of the position and scale of the \emph{unknown} shapes in the images of the testing set. The mean is computed over all the 218 images in this set.}
\label{tab:ExpTestingDataset}
\begin{tabular*}{\columnwidth}{@{\extracolsep{\fill}} | c | c | c | c | c |} \hline 
$\overline{|\mathbb{A}|}$   &   $\overline{\mu_{\vec{t}}}$ (pixels) & $\overline{\sigma_{\vec{t}}}$ (pixels) &  $\overline{\mu_S}$ (\%) & $\overline{\sigma_S}$ (\%)  \\ \hline 
271 & 1.36 & 2.26 & 2.70 & 6.64 \\ \hline
 \end{tabular*}
\vspace{-10pt}
\end{table}

Note in Table \ref{tab:ExpTestingDataset} that the average number of solutions per image (271) is only a small fraction (0.06\%) of the total number of hypotheses (471,900). Moreover, these solutions are (in general) concentrated near the ``ground truth" defined above (judging from the mean ``errors" in the table).

\begin{figure}
\vspace{-5pt}
\begin{center}
\includegraphics[width=\columnwidth, bb=0pt 0pt 1190pt 793pt]{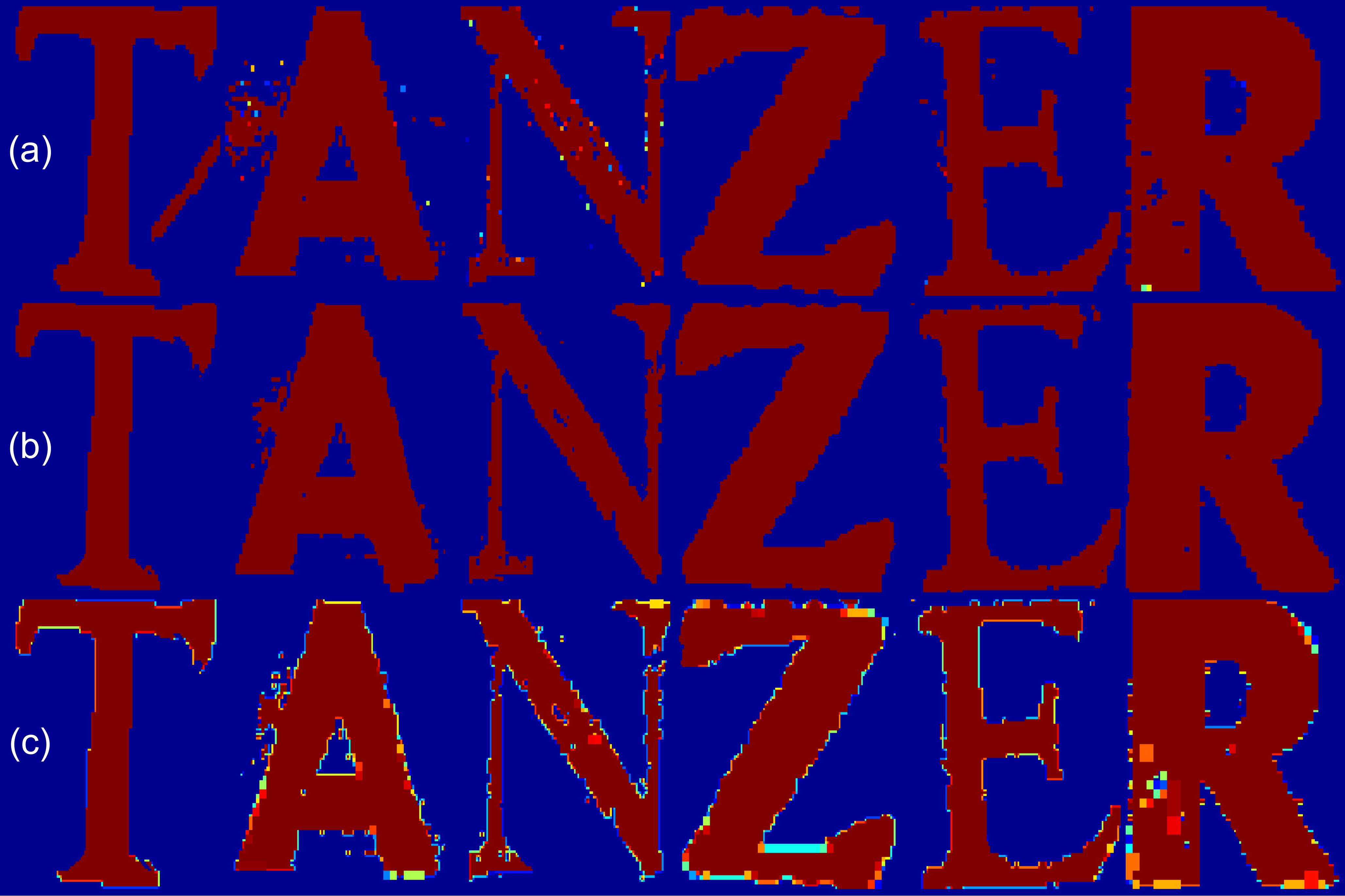}
\end{center}
\vspace{-10pt}
\caption{Examples of segmentations obtained in this experiment. (a) The original BFs computed from the input images. (b) Discrete shapes estimated for each of the BFs in (a). (c) Semidiscrete shapes estimated for each of the BFs in (a). As in Fig. \ref{fig:Exp3Noise}, the colors in these semidiscrete shapes represent the \emph{fraction} of each partition element that is ``covered" by the shape.}
\label{fig:RealExpSegmentations}
\vspace{-10pt}
\end{figure}

To illustrate the types of segmentations obtained by the framework, Fig. \ref{fig:RealExpSegmentations} shows the shapes estimated for some of the images in the testing set. Note how most of the ``noise" has been eliminated.

In order to understand why misclassifications occurr, let us look at some solutions found by the proposed approach. Each column in Fig. \ref{fig:TypicalResults} shows two solutions found for a single testing image. In the top row a solution of the correct class is depicted, while in the bottom row a solution from a different class is depicted. It can be seen in the figure that when the fit of the correct solution is not very good, solutions of an incorrect class cannot be discarded. This happens in the case of rotated or distorted letters (e.g., `D' and `N' in the figure) and in cases where there is not a good fit in the database of priors (e.g., `F' in the figure). These erros can possibly be overcome by considering richer transformations and/or more numerous priors, respectively (more on this in the next section).

\section{Conclusions}
\label{sec:Conclusions}
This article presented a new type of algorithms, namely hypothesize-and-bound algorithms, to significantly trim down the amount of computation required to select the hypothesis that best explains an input (e.g., an image), from a set of predefined hypotheses. These algorithms are specifically targeted to perceptual problems in which the large size of the input imposes that these problems be solved by ``paying attention'' to a small fraction of the input only. 

These algorithms have two important properties: 1) they are \emph{guaranteed to find the set of optimal hypotheses}; and 2) \emph{the total amount of computation required by them is decoupled from the ``resolution'' of the input, and only depends on the current problem}. 

After describing the general paradigm of H\&B algorithms, we instantiated this paradigm to the problem of simultaneously estimating the class, pose, and a denoised version of a 2D shape in a 2D image, by incorporating prior knowledge about the possible shapes that can be found in the image. In order to instantiate the paradigm, we developed a novel theory of shapes and shape priors that allowed us to develop the BM for this particular problem. 

We consider that the theory and algorithms proposed here are not just limited to solve the problem described in this article, but rather that they are general enough as to find application in many other domains. In fact, the application of the exact same FoAM (described in Section \ref{sec:FoAM}) and theory of shapes (described in Section \ref{sec:TheoryOfShapes}) have already been instantiated to solve the problem of simultaneously estimating the class, pose, and a 3D reconstruction of an object from a single 2D image \cite{Rot102}.

Even though the contributions of this work are main\-ly theoretical, we showed preliminary results of the proposed framework on a real problem. Nevertheless, we acknowledge that to provide state-of-the-art results the framework needs to be extended in several directions (which is beyond the scope of the current article). For example, a limitation of the current framework is in its ability to deal with classes of shapes that are too diverse. These classes result in priors with large parts whose probabilities are not close to 0 or 1, and thus the bounds obtained are not tight, reducing the efficacy of the proposed method. These \emph{unspecific} priors are obtained, for example, when the parts of the shapes in the class are not in consistent positions. To address this problem, in Section \ref{sec:RealExperiments}, we divided highly variable classes into subclasses with less variability. This addressed the problem created by the highly variable classes, at the expense of creating many more hypotheses.

Another direction to extend this work that addresses the growing number of hypotheses mentioned above, is to exploit the redundancy among these hypotheses. It was explained in Section \ref{sec:SyntheticExperiments} that to prove that two (or more) hypotheses are indistinguishable, it is necessary to refine these hypotheses completely. By doing so, however, the performance drops significantly. This problem gets exacerbated as the number of indistinguishable hypotheses increases, which in turn happens when classes are divided in subclasses or when the number of degrees of freedom are increased. In order to exploit the redundancy among hypotheses, bounds can be computed not just for individual hypotheses but for groups of them, \emph{a la} Branch \& Bound. This new algorithm would then work by splitting not just the image domain (as H\&B algorithms do), but also the space of hypotheses (as Branch \& Bound algorithms do). This new algorithm is expected to be most effective when the hypotheses are very similar, which is exactly the case that the current approach is less efficient dealing with. \textcolor{white}{\cite{Sch00}.}

\begin{acknowledgements}
We would like to thank Siddharth Mahendran for his help running part of the experiments.
\end{acknowledgements}

\clearpage
\bibliographystyle{spmpsci}      
\bibliography{Rother}   

\clearpage
\section{Supplementary material}
\label{sec:SupplementaryMaterial}

\setcounter{lemma}{1}
\addtocounter{lemma}{-1}
\begin{lemma}[Properties of LCDFs and \emph{m}-Su\-mma\-ries]
Let $\Pi$ be an arbitrary partition, and let $\tilde{S}$ and $D={\left\{D_{\omega}\right\}}_{\omega \in \Pi}$ be respectively a semidiscrete shape and a LCDF in this partition. Then:

1. Given a BF $B$ such that $B\sim D$, it holds that (Fig. \ref{fig:LCDFs})
\begin{equation}
\tag{\ref{eq:LemaPart1}}
\mathop{\sup}_{S\sim \tilde{S}} \int_{\Omega}{\delta_B(\vec{x})S(\vec{x})\ d\vec{x}} = 
\sum_{\omega \in \Pi}{\int^{\left|\omega\right|}_{\left|\omega\right|-\tilde{S}(\omega)} {D^{-1}_{\omega}(a)\ da}}.
\end{equation}

2. Similarly, given a continuous shape $S$, such that $S\sim \tilde{S}$, it holds that
\begin{equation}
\tag{\ref{eq:LemaPart2}}
\mathop{\sup}_{B\sim D} \int_{\Omega}{\delta_B(\vec{x})S(\vec{x})\ d\vec{x}} =
\sum_{\omega \in \Pi} {\int^{\left|\omega\right|}_{\left|\omega\right|-\tilde{S}(\omega)} {D^{-1}_{\omega}(a)\ da}}.
\end{equation}

3. Moreover, for any $\alpha \in \left[0, |\omega|\right]$, the integrals on the rhs of \eqref{eq:LemaPart1} and \eqref{eq:LemaPart2} can be bounded as
\begin{align}
\int^{\left|\omega\right|}_{\alpha}{D^{-1}_{\omega}(a)\ da} \le \frac{\delta_{max}}{m} \left[ 
\vphantom{\sum^{m-1}_{j=J(\alpha)}{\left(\tilde{Y}^{j+1}_{\omega}\right)(j+1)}}
J(\alpha)\left(\tilde{Y}^{J(\alpha)}_{\omega}-\alpha \right) + \right. \nonumber \\
\tag{\ref{eq:LemaPart3}}
\left. \sum^{m-1}_{j=J(\alpha)}{\left(\tilde{Y}^{j+1}_{\omega}-\tilde{Y}^j_{\omega}\right)(j+1)} \right],
\end{align}
where $J(\alpha)$ is as defined in \eqref{eq:J}.

4. The rhs of \eqref{eq:LemaPart1} and \eqref{eq:LemaPart2} are maximum when $\tilde{S}(\omega)=|\omega|-A_{\omega}$, thus
\begin{equation}
\tag{\ref{eq:LemaPart4}}
\mathop{\sup}_{\tilde{S}} \sum_{\omega \in \Pi} {\int^{|\omega|}_{|\omega|-\tilde{S}(\omega)}{D^{-1}_{\omega}(a)\ da}} = 
\sum_{\omega \in \Pi} {\int^{|\omega|}_{A_{\omega}}{D^{-1}_{\omega}(a)\ da}}.
\end{equation}
\begin{proof}
\hspace{-4pt}: 1. For simplicity, let us only consider the case in which the partition $\Pi$ consists of a single element $\omega$; the generalization of the proof to partitions containing multiple elements is straightforward. Let us assume for the moment that $D_{\omega}(\delta)$ is strictly increasing and continuous, so that $D_{\omega}\left(D_{\omega}^{-1}(a) \right) = a \ \forall a \in [0, |\omega|]$.

Consider the continuous shape $S^*$, defined as 
\begin{equation}
\label{eq:L1SStar}
S^*(\vec{x}) \triangleq \mathit{1} \left( \delta_B(\vec{x}) - \tilde{\delta} \right) = \left\{
\begin{array}{ll}
1, & \ \text{if} \ \delta_B(\vec{x}) \ge \tilde{\delta} \\
0, & \ \text{otherwise},
\end{array}
\right.
\end{equation}
where $\mathit{1}(\cdot)$ is the Heaviside step function and $\tilde{\delta} \triangleq D_{\omega}^{-1}$ $\left(\left|\omega\right|-\tilde{S}(\omega) \right)$. This shape is the solution to the lhs of \eqref{eq:LemaPart1}. To see this, notice that $S^* \sim \tilde{S}$ because
\begin{align}
\int_{\omega}{S^*(\vec{x})} \ d\vec{x} & = \left| \left\{ \vec{x} \in \omega: \delta_B(\vec{x}) \ge \tilde{\delta}\right\} \right| \nonumber \\
& = \left| \omega \right| - D_{\omega}\big(\tilde{\delta}\big) = \tilde{S}(\omega).
\end{align}
Notice also that $S^*$ maximizes the integral on the lhs of \eqref{eq:LemaPart1} because it contains the parts of $\omega$ with the highest values of $\delta_B$.

Thus, substituting \eqref{eq:L1SStar} on the lhs of \eqref{eq:LemaPart1}, and defining the function $g(\vec{x}) \triangleq \delta_B(\vec{x}) - \tilde{\delta}$, \eqref{eq:LemaPart1} reduces to
\begin{align}
\int_{\omega}{\delta_B(\vec{x}) \mathit{1}\left( \delta_B(\vec{x}) - \tilde{\delta} \right) \ d\vec{x}} & = \nonumber \\ 
\int_{\omega}{ g(\vec{x}) \mathit{1}\left(g(\vec{x})\right) \ d\vec{x}} 
+ \tilde{\delta} \int_{\omega}{ \mathit{1}\left(g(\vec{x})\right) \ d\vec{x}} &.
\label{eq:L1Step2}
\end{align}

Using Lebesgue integration \cite{Wil78}, this last expression is equivalent to
\begin{align}
\int_0^{\infty}{\left| \left\{ 
\vec{x} \in \omega: g(\vec{x}) \ge \delta
\right\} \right| \ d\delta} + \nonumber \\
\tilde{\delta} \left| \left\{ 
\vec{x} \in \omega: g(\vec{x}) \ge 0
\right\} \right| = \nonumber \\
\int_0^{\infty}{\left| \left\{ 
\vec{x} \in \omega: \delta_B(\vec{x}) \ge \delta + \tilde{\delta}
\right\} \right| \ d\delta} + \nonumber \\
 \tilde{\delta} \left| \left\{ 
\vec{x} \in \omega: \delta_B(\vec{x}) \ge \tilde{\delta}
\right\} \right| = \nonumber \\
\int_0^{\infty}{\left[|\omega| - D_{\omega}(\delta + \tilde{\delta}) \right] \ d\delta} + 
\tilde{\delta} \left(|\omega| - D_{\omega}(\tilde{\delta}) \right) = \nonumber \\
\int_{ \tilde{\delta}}^{\delta_{max}}{\left[|\omega| - D_{\omega}(\delta) \right] \ d\delta} + 
\tilde{\delta} \tilde{S}(\omega).
\label{eq:L1Step3}
\end{align}

Now recall that the integral of a function $f(x)$ can be written in terms of its inverse $f^{-1}(y)$, if it exists, as \cite{Sch00}
\begin{equation}
\int_a^b{f(x) \ dx}=x f(x)|_a^b - \int_{f(a)}^{f(b)}{f^{-1}(y) \ dy}.
\end{equation}
Hence, it follows that \eqref{eq:L1Step3} is equal to
\begin{align}
|\omega| & (\delta_{max} - \tilde{\delta}) - \delta_{max} D_{\omega}(\delta_{max}) + \tilde{\delta} D_{\omega}(\tilde{\delta}) + \nonumber \\
& \int_{D_{\omega}(\tilde{\delta})}^{D_{\omega}(\delta_{max})}{D_{\omega}^{-1}(a) \ da} + 
\tilde{\delta} \tilde{S}(\omega) = \nonumber \\
|\omega| & (\delta_{max} - \tilde{\delta}) - \delta_{max} |\omega| + \tilde{\delta} (|\omega| - \tilde{S}(\omega)) + \nonumber \\
& \int_{|\omega| - \tilde{S}(\omega)}^{|\omega|}{D_{\omega}^{-1}(a) \ da} + 
\tilde{\delta} \tilde{S}(\omega) = \nonumber \\
& \int_{|\omega| - \tilde{S}(\omega)}^{|\omega|}{D_{\omega}^{-1}(a) \ da},
\end{align}
as we wanted to prove. 

If $D_{\omega}(\delta)$ is not one-to-one (as we assumed above), the proof is essentially similar but complicated by the fact that $S^*$ can be chosen in more than one way (because there are sets $\theta \subset \omega$, with $|\theta|>0$, where $\delta_B$ is constant) and this has to be handled explicitly. \qed

2. To prove 2 we proceed exactly as in 1, except that in this case we choose $\delta_B$ to have its largest values in $S$, instead of choosing $S$ to be supported where the largest values of $\delta_B$ are. \qed

3. From \eqref{eq:DInverseUpper}, it follows that
\begin{align}
\int^{|\omega|}_{\alpha}{D^{-1}_{\omega}(a)\ da} \le
\frac{\delta_{max}}{m} \int^{|\omega|}_{\alpha}{J(a)\ da} = \nonumber \\
\label{eq:LemaPart3Proof}
\frac{\delta_{max}}{m}\left[\int^{\tilde{Y}^{J(\alpha)}_{\omega}}_{\alpha}{J(\alpha)\ da} + \sum^{m-1}_{j=J(\alpha)}{\int^{\tilde{Y}^{j+1}_{\omega}}_{\tilde{Y}^j_{\omega}}{(j+1)\ da}}\right].
\end{align}
And because the integrands on the rhs of \eqref{eq:LemaPart3Proof} are all constant, \eqref{eq:LemaPart3} follows. \qed

4. It is clear that each integral on the lhs of \eqref{eq:LemaPart4} is maximum when the integration domain contains only the parts where $D^{-1}_{B,\omega}(a)$ is positive or zero. Therefore, from \eqref{eq:XCrossingPoint}, each integral is maximum when computed in the interval $\left[A_{B,\omega},|\omega|\right]$, which yields the rhs of \eqref{eq:LemaPart4}, proving the lemma. \qed
\end{proof}
\end{lemma}

\setcounter{theorem}{1}
\addtocounter{theorem}{-1}
\begin{theorem}[Lower bound for $L(H)$]
Let $\Pi$ be a partition, and let $\hat{Y}_f={\left\{\hat{Y}_{f,\omega}\right\}}_{\omega \in \Pi}$ and $\hat{Y}_H={\left\{\hat{Y}_{H,\omega}\right\}}_{\omega \in \Pi}$ be the \emph{mean}-summaries of two unknown BFs in $\Pi$. Then, for any $B_f \sim \hat{Y}_f$ and any $B_H \sim \hat{Y}_H$, it holds that $L(H) \ge \underline{L}_{\Pi}(H)$, where
\begin{align}
\tag{\ref{eq:LowerBound}}
\underline{L}_{\Pi}(H) & \triangleq Z_{B_H} + \sum_{\omega \in \Pi}{\underline{\mathcal L}_{\omega}(H)}, \\
\tag{\ref{eq:LowerBoundLocal}}
\underline{\mathcal L}_{\omega}(H) & \triangleq \left(\hat{Y}_{f,\omega}+\hat{Y}_{H,\omega}\right)\hat{q}_*(\omega),
\end{align}
and $\hat{q}_*$ is a discrete shape in $\Pi$ defined as
\begin{equation}
\tag{\ref{eq:DiscreteShapeLower}}
\hat{q}_*(\omega) \triangleq \left\{ \begin{array}{ll}
	1, & \ \text{if} \left(\hat{Y}_{f,\omega} + \hat{Y}_{H,\omega}\right)>0 \\ 
	0, & \ \text{otherwise}. \end{array}
\right.
\end{equation}
\begin{proof}
\hspace{-4pt}: Since the set of continuous shapes $q$ that are compatible with the discrete shape $\hat{q}$ is a subset of the set of all continuous shapes (from \eqref{eq:DiscreteShapesInContinuousShapes}), it holds that
\begin{align}
L(H) & \ge Z_{B_H} + \mathop{\max}_{\hat{q}} \left[\mathop{\sup}_{q \sim \hat{q}} \int_{\Omega} {\left(\delta_{B_f}(\vec{x}) + \delta_{B_H}(\vec{x})\right)q(\vec{x})\ d\vec{x}} \right] \\
\label{eq:LBDStep1}
& = Z_{B_H} + \mathop{\max}_{\hat{q}} {\sum_{\omega \in \Pi}{\hat{q}(\omega) \int_{\omega}{\left(\delta_{B_f}(\vec{x}) + \delta_{B_H}(\vec{x}) \right)\ d\vec{x}} }}.
\end{align}

The terms in the integral in \eqref{eq:LBDStep1} are, by definition, the mean-summaries of the BFs (see \eqref{eq:MeanSummary}), thus
\begin{equation}
\label{eq:LBDStep2}
L(H) \ge Z_{B_H} + \mathop{\max}_{\hat{q}} \sum_{\omega \in \Pi}{\hat{q}(\omega) \left(\hat{Y}_{f,\omega} + \hat{Y}_{H,\omega} \right)}.
\end{equation}

It then follows that the discrete shape defined in \eqref{eq:DiscreteShapeLower} maximizes the rhs of \eqref{eq:LBDStep2}, proving the theorem. \qed
\end{proof}
\end{theorem}

\setcounter{theorem}{2}
\addtocounter{theorem}{-1}
\begin{theorem}[Upper bound for $L(H)$]
Let $\Pi$ be a partition, and let $\tilde{Y}_f={\left\{\tilde{Y}_{f,\omega}\right\}}_{\omega \in \Pi}$ and $\tilde{Y}_H={\left\{\tilde{Y}_{H,\omega}\right\}}_{\omega \in \Pi}$ be the \emph{m}-summaries of two unknown BFs in $\Pi$. Let $\tilde{Y}_{f \bigoplus H,\omega}$ ($\omega \in \Pi$) be a vector of length $4m+2$ obtained by sorting the values in $\tilde{Y}_{f,\omega}$ and $\tilde{Y}_{H,\omega}$ (in ascending order), keeping repeated values, i.e.,
\begin{align}
\tilde{Y}_{f \bigoplus H,\omega} & \triangleq \left[\tilde{Y}^1_{f \bigoplus H,\omega},\dots,\tilde{Y}^{4m+2}_{f \bigoplus H,\omega} \right] \nonumber \\
\tag{\ref{eq:SortedSummaries}}
& \triangleq SortAscending\left(\tilde{Y}_{f,\omega} \cup \tilde{Y}_{H,\omega}\right).
\end{align}

Then, for any $B_f \sim \tilde{Y}_f$ and any $B_H \sim \tilde{Y}_H$, it holds that $L(H) \le \overline{L}_{\Pi}(H)$, where 
\begin{align}
\tag{\ref{eq:UpperBound}}
\overline{L}_{\Pi}(H) & \triangleq Z_{B_H} + \sum_{\omega \in \Pi} {\overline{\mathcal L}_{\omega}(H)}, \ \text{and} \\
\tag{\ref{eq:UpperBoundLocal}}
\overline{\mathcal L}_{\omega}(H) & \triangleq \frac{\delta_{max}}{m} \sum_{j=2m+1}^{4m+1}{(j-2m) \left( \tilde{Y}^{j+1}_{f \bigoplus H,\omega} - \tilde{Y}^j_{f \bigoplus H,\omega} \right)}.
\end{align}
It also follows that the continuous shape that maximizes \eqref{eq:L_H} is equivalent to a semidiscrete shape $\tilde{q}_*$ in $\Pi$ that satisfies
\begin{equation}
\tag{\ref{eq:SemiDiscreteShapeUpper}}
\tilde{q}_*(\omega) \in \left[|\omega| - \tilde{Y}^{2m+1}_{f \bigoplus H,\omega},|\omega| - \tilde{Y}^{2m}_{f \bigoplus H,\omega}\right] \ \forall \omega \in \Pi.
\end{equation}

\begin{proof}
\hspace{-4pt}: Let $D_f$ and $D_H$ be two arbitrary cumulative distributions such that $D_f \sim \tilde{Y}_f$ and $D_H \sim \tilde{Y}_H$. Then, 
\begin{align}
L(H) \le Z_{B_H} + \nonumber \\
\mathop{\sup}_{
\begin{array}{c} 
\scriptstyle{B'_f \sim D_f} \\ 
\scriptstyle{B'_H\sim D_H} 
\end{array}}
\mathop{\sup}_{q} \int_{\Omega}{q(\vec{x})\left(\delta_{B'_f}(\vec{x})+\delta_{B'_H}(\vec{x})\right)\ d\vec{x}},
\end{align}
and from \eqref{eq:SemiDiscreteShapesEqualContinuousShapes}, this expression is equal to
\begin{align}
&= Z_{B_H} + \mathop{\sup}_{
\begin{array}{c}
\scriptstyle{B'_f \sim D_f} \\
\scriptstyle{B'_H\sim D_H} 
\end{array}} 
\mathop{\sup}_{\tilde{q} \in \widetilde{\mathbb S}(\Pi)} \nonumber \\
& \mathop{\sup}_{q \sim \tilde{q}} \int_{\Omega} { \left( \delta_{B'_f}(\vec{x}) + \delta_{B'_H}(\vec{x}) \right) q(\vec{x})\ d\vec{x}}.
\label{eq:UBDStep1}
\end{align}

Exchanging the order of the \emph{sup} operations and using \eqref{eq:LemaPart2}, the rhs of \eqref{eq:UBDStep1} is less or equal than
\begin{align}
Z_{B_H} + \mathop{\sup}_{\tilde{q} \in \widetilde{\mathbb S}(\Pi)} \left[\mathop{\sup}_{q \sim \tilde{q}} \left[ \mathop{\sup}_{B'_f \sim D_f} \int_{\Omega}{{\delta_{B'_f}(\vec{x}) q(\vec{x})\ d\vec{x}} } + \right.\right. \nonumber \\
\left.\left. \mathop{\sup}_{B'_H \sim D_H} \int_{\Omega}{\delta_{B'_H}(\vec{x}) q(\vec{x})\ d\vec{x}} \right]\right]= \nonumber
\end{align}
\begin{align}
Z_{B_H} + \mathop{\sup}_{\tilde{q} \in \widetilde{\mathbb S}(\Pi)} \left[ \sum_{\omega \in \Pi}
\int^{|\omega|}_{|\omega|-\tilde{q}(\omega)} {D^{-1}_{f,\omega}(a)\ da} + \right. \nonumber \\
\left. \int^{|\omega|}_{|\omega|-\tilde{q}(\omega)} {D^{-1}_{H,\omega}(a)\ da} \right]= \nonumber \\
Z_{B_H} + \sum_{\omega \in \Pi} \mathop{\sup}_{\tilde{q}(\omega)}
\int^{|\omega|}_{|\omega|-\tilde{q}(\omega)} {\left(D^{-1}_{f,\omega}(a) + D^{-1}_{H,\omega}(a)\right)\ da}.
\label{eq:UBD:Step2}
\end{align}

Since $D_f \sim \tilde{Y}_f$ and $D_H \sim \tilde{Y}_H$, it follows from \eqref{eq:DInverseUpper} and \eqref{eq:J} that 
\begin{align}
D^{-1}_{f,\omega}(a) + D^{-1}_{H,\omega}(a) \le \nonumber \\
\frac{\delta_{max}}{m} \left(
\left|\left\{j:\tilde{Y}^j_{f,\omega}<a\right\}\right| +
\left|\left\{j:\tilde{Y}^j_{H,\omega}<a\right\}\right|-2m
\right),
\end{align}
which using \eqref{eq:SortedSummaries} can be rewritten as
\begin{equation}
\overline{D^{-1}_{f \bigoplus H,\omega}} (a) \triangleq 
\frac{\delta_{max}}{m} \left(\left|\left\{j:\tilde{Y}^j_{f \bigoplus H,\omega}<a\right\}\right|-2m\right). 
\end{equation}

Therefore, it follows that
\begin{equation}
\nonumber
\int^{|\omega|}_{\tilde{Y}^k_{f \bigoplus H,\omega}} {\overline{D^{-1}_{f \bigoplus H,\omega}}(a)\ da}=
\end{equation}
\begin{equation}
\nonumber
=\frac{\delta_{max}}{m} \sum^{4m+1}_{j=k} {\int^{\tilde{Y}^{j+1}_{f \bigoplus H,\omega}}_{\tilde{Y}^j_{f \bigoplus H,\omega}}{(j-2m)\ da}}=
\end{equation}
\begin{equation}
\label{eq:UBDStep3}
=\frac{\delta_{max}}{m} \sum^{4m+1}_{j=k}{(j-2m) \left(\tilde{Y}^{j+1}_{f \bigoplus H,\omega} - \tilde{Y}^j_{f \bigoplus H,\omega} \right)}.
\end{equation}

Substituting this last expression into \eqref{eq:UBD:Step2} yields the bound
\begin{equation}
\label{eq:UBDStep4}
\begin{split}
L(H) \le Z_{B_H} + \frac{\delta_{max}}{m} \sum_{\omega \in \Pi} \mathop{\sup}_{k \in \left\{1,\dots,4m+1\right\}}\\
 \sum^{4m+1}_{j=k}{(j-2m) \left(
\tilde{Y}^{j+1}_{f \bigoplus H,\omega} - 
\tilde{Y}^j_{f \bigoplus H,\omega}
\right)}.
\end{split}
\end{equation}

Since $\overline{D^{-1}_{f \bigoplus H,\omega}}(a)$ is a non-decreasing, piecewise constant function that takes the zero value when $a \in ($ $\tilde{Y}^{2m}_{f \bigoplus H,\omega},$ $\tilde{Y}^{2m+1}_{f \bigoplus H,\omega}]$, it follows that the semidiscrete shape in \eqref{eq:SemiDiscreteShapeUpper} maximizes \eqref{eq:UBD:Step2}, and that the supremum in \eqref{eq:UBDStep4} is obtained when $k=2m+1$, proving the theorem. \qed
\end{proof}
\end{theorem}

\end{document}